\documentclass[acmsmall]{acmart}
\usepackage{circuitikz}

\usepackage{multirow}
\usepackage{textcomp}
\usepackage{hyperref}
\usepackage[normalem]{ulem}
\useunder{\uline}{\ul}{}

\AtBeginDocument{%
  \providecommand\BibTeX{{%
    \normalfont B\kern-0.5em{\scshape i\kern-0.25em b}\kern-0.8em\TeX}}}

\setcopyright{rightsretained}
\copyrightyear{2024}
\acmYear{2024}
\acmDOI{XXXXXXX.XXXXXXX}

\acmJournal{JACM}
\acmVolume{37}
\acmNumber{4}
\acmArticle{111}
\acmMonth{8}

\usepackage{amsmath}
\usepackage{wrapfig}
\usepackage{pifont}
\usepackage[symbol]{footmisc}

\begin{document}

\title[Security and Privacy Challenges of Large Language Models: A Survey]{Security and Privacy Challenges of Large Language Models:\\ A Survey}

\author{Badhan Chandra Das}
\affiliation{%
  \institution{Knight Foundation School of Computing and Information Sciences; Sustainability, Optimization, and Learning for InterDependent networks laboratory (solid lab), Florida International University}
  \city{Miami}
   \state{Florida}
  \country{United States}}

\author{M. Hadi Amini}
\authornote{Corresponding authors.}

\affiliation{%
  \institution{Knight Foundation School of Computing and Information Science, solid lab, Florida International University}
  \city{Miami}
  \state{Florida}
  \country{United States}}

\author{Yanzhao Wu}
\authornotemark[1]

\affiliation{
  \institution{Knight Foundation School of Computing and Information Sciences, Florida International University}
  \city{Miami}
   \state{Florida}
  \country{United States}}
  \email{Emails: {moamini,yawu}@fiu.edu}

\renewcommand{\shortauthors}{Das, et al.}

\begin{abstract}

Large language models (LLMs) have demonstrated extraordinary capabilities and contributed to multiple fields, such as generating and summarizing text, language translation, and question-answering. Nowadays, LLMs have become very popular tool in natural language processing (NLP) tasks, with the capability to analyze complicated linguistic patterns and provide relevant and appropriate responses depending on the context.
While offering significant advantages, these models are also vulnerable to security and privacy attacks, such as jailbreaking attacks, data poisoning attacks, and personally identifiable information (PII) leakage attacks. This survey provides a thorough review of the security and privacy challenges of LLMs, along with the application-based risks in various domains, such as transportation, education, and healthcare. 
We assess the extent of LLM vulnerabilities, investigate emerging security and privacy attacks for LLMs, and review the potential defense mechanisms. Additionally, the survey outlines existing research gaps in this research area and highlights future research directions.

\end{abstract}



\maketitle

\section{Introduction}

The exploration of intelligence and the feasibility of machines with cognitive abilities is a compelling pursuit in the scientific community. Intelligent devices equip us with the capacity for logical reasoning, experimental inquiry, and foresight into future developments. In the Artificial Intelligence (AI) domain, researchers are diligently striving to advance methodologies for the construction of intelligent machines. One of the latest advancements of AI is LLM. LLMs have become popular in both the academic and industrial sectors. As researchers demonstrate, these models are impressively effective and achieve nearly human-like performance in certain tasks \cite{zhuang2023efficiently}. Consequently, there is growing interest in exploring whether they might represent an early form of Artificial General Intelligence (AGI). Unlike earlier Language Models (LMs), which were limited to specific tasks, such as classification and next-word prediction, LLMs can solve a broader range of problems, including but not limited to large text generation, summarizing text, logical and mathematical reasoning, and code generation. They are highly capable of handling various tasks, from daily use of language for communication to more specific challenges \cite{chung2024scaling}, \cite{jiang2023lion}, \cite{jin2023rethinking}, \cite{raffel2020exploring}. 

\begin{wraptable}{r}{0.6\textwidth}
\centering
\scalebox{.7}{
\begin{tabular}{|c|l|}

\hline
\textbf{Acronym} & \multicolumn{1}{c|}{\textbf{Full Form}}                         \\ \hline
AI               & Artificial Intelligence                                         \\ \hline
AGI              & Artificial General Intelligence                                 \\ \hline
ALBERT           & A Lite BERT                                                     \\ \hline
BERT             & Bidirectional Encoder Representations from Transformers         \\ \hline
BGMAttack        & Black-box Generative Model-based Attack                          \\ \hline
CBA              & Composite Backdoor Attack                                       \\ \hline

CCPA             & California Consumer Privacy Act                                 \\ \hline
DAN              & Do Anything Now                                                 \\ \hline
DNN              & Deep Neural Network                                             \\ \hline
DP               & Differential Privacy                                            \\ \hline
FL               & Federated Learning                                               \\ \hline
GDPR             & General Data Protection Regulation                              \\ \hline
GA               & Genetic Algorithm                                               \\ \hline
GPT               & Generative Pre-trained Transformer                             \\ \hline

HIPAA            & Health Insurance Portability and Accountability Act             \\ \hline
LM               & Language Model                                                  \\ \hline
LLM              & Large Language Model                                            \\ \hline
Llama            & Large Language Model Meta AI                                    \\ \hline
MIA              & Membership Inference Attack                                     \\ \hline
MDP              & Masking-Differential Prompting                                  \\ \hline
MLM              & Masked Language Model                                           \\ \hline

NLP              & Natural Language Processing                                     \\ \hline

OOD              & Out Of Distribution                                             \\ \hline
PI               & Prompt Injection                                                \\ \hline

PII              & Personally Identifiable Information                             \\ \hline

PAIR             & Prompt Automatic Iterative Refinement                           \\ \hline
PLM              & pre-trained Language Model                                      \\ \hline
RL               & Reinforcement Learning                                      \\ \hline
RLHF             & Reinforcement Learning from Human Feedback                      \\ \hline
RoBERTa          & Robustly optimized BERT approach                                \\ \hline
SGD              & Stochastic Gradient Descent                                     \\ \hline

TAG              & Gradient Attack on Transformer-based Language Models            \\ \hline

XLNet            & Transformer-XL with autoregressive and autoencoding pre-training \\ \hline
\end{tabular}
}

 \caption{List of Common Acronyms}
    
    \label{tab:acr}
  
\end{wraptable}

Also, with proper prompt engineering \cite{wei2022chain} and in-context learning capabilities \cite{kojima2022large}, LLMs can adapt to different contexts and/or even accomplish new tasks without training or fine-tuning. The introduction of ChatGPT \cite{chatGPT} and GPT-4 \cite{GPT4} took these advancements to another level. However, these highly efficient LLMs are not flawless. The vulnerabilities of these LLMs have not been explored that much on a large scale from security and privacy perspective. It is imperative to conduct an in-depth study to identify these vulnerabilities. In this paper, we comprehensively illustrate the security and privacy issues in LLMs as well as their defense mechanisms. We also discuss the research challenges in the LLM context along with future research opportunities.

Throughout the paper, there are many acronyms used to represent concepts, types of attacks, models common in privacy and security, and LLM research very frequently. Table \ref{tab:acr} is provided for the most common and important terms we used in the paper.

\subsection{Motivation}
The increasing sizes of LMs, such as LLMs, require a huge amount of data from the Internet in addition to meticulously annotated textual data for training/fine-tuning to enhance models' predictive performance. In contrast to carefully created annotated data, the freely available texts from the Internet may exhibit poor data quality and unintended leakage of private personal information \cite{li2023privacy}. For instance, casual interactions with these models may accidentally leak PII, as highlighted in \cite{carlini2021extracting} and \cite{li2023multi}, which may violate existing privacy laws, such as The ``Health Insurance Portability and Accountability Act of 1996 (HIPAA)'' in the United States \cite{cheng2006health}, the EU's ``General Data Protection Regulation (GDPR)'' \cite{voigt2017eu}, and the ``California Consumer Privacy Act (CCPA)'' \cite{CCPA}.

\begin{table}[!ht]
\centering
\small
\scalebox{.65}{
{\color{black}
                              & Survey                                                          & $\large***$                                                                   & $\large***$                                                  & $\large***$                                                 & $\large***$                                               & $\large***$                                                        & $\large***$                                                           & $\large***$                                                              & $\large***$                                   &    $\large***$           &         $\large***$                                      & $\large***$            &               Current                           \\ \hline
\end{tabular}
}
}

\caption{Comparison of the Existing Surveys and Research Works on LLM Vulnerabilities with our paper. The acronyms stands as JA: Jailbreaking Attack, PI: Prompt Injection, BA: Backdoor Attack, DPa: Data Poisoning Attack, GLa: Gradient Leakage Attack, MIA: Membership Inference Attack, PII-Leak: Personal Identifiable Information Leakage Attacks, PD: Publication Date, 1-1 DAA: one-to-one Defense Against Attacks, AFRD: Analysis and Future Research Direction. We define the extent of discussion by notations as: No Discussion($\times$), Slight Discussion({\large$*$}), Moderate Discussion({\large$**$}), Extensive Discussion({\large$***$)}}
\label{tab:comp}
 
\end{table}

Following the launch of ChatGPT \cite{chatGPT} and GPT-4 \cite{GPT4}, numerous research initiatives have focused on assessing them across various dimensions. These evaluations considered various aspects of NLP tasks, such as correctness, robustness, rationality, reliability, and notably, the identification and evaluation of vulnerabilities related to privacy risks and security issues. The assessment of LLMs is of paramount importance for several reasons. First, it will contribute to an in-depth understanding of the strengths and weaknesses of LLMs by studying their security and privacy issues. Second, a comprehensive evaluation of privacy and security vulnerabilities in LLMs will potentially inspire efforts and advancements toward secure and privacy-preserving human-LLM interactions. Third, the widespread use of LLMs highlights the significance of assuring their reliability and security, particularly in sectors prioritizing safety and privacy protection, such as financial organizations and the healthcare system. Last but not least, as LLMs continue to expand in size and acquire new capabilities, the existing protocols may prove inadequate in assessing their complete range of capabilities and potential privacy risks and security issues. Our objective is to provide a clear vision for researchers, practitioners, and other stakeholders who plan to develop and/or deploy LLMs regarding the significance of LLM security and privacy challenges. This involves reviewing existing studies in the broad area of security, privacy, and their intersections, and notably, highlighting future research directions to design novel evaluation protocols and attack methods, as well as defense mechanisms tailored to the evolving landscape of LLMs.

\subsection{Our Contributions}
This paper analyzes the latest developments in privacy and security concerns and defense mechanisms of LLMs. Comparing with recent survey papers and empirical studies on this topic as shown in Table \ref{tab:comp}, 
we present a comprehensive discussion and systematic analysis of representative privacy and security issues, defense mechanisms, and future research directions for LLMs. 
In contrast to the prior surveys, we investigated the most recent advancements in the security and privacy domain for LLMs, providing a timely and highly relevant review of this emerging research area. Furthermore, our study analyzed novel approaches and techniques that emerged in this domain and the current research gaps. After analyzing the effectiveness and limitations of representative attacks and defenses, we offer insights into future research directions on unexplored security and privacy challenges and potential attack mitigation strategies.

\begin{figure}[!ht]
    \centering
    \includegraphics[width=1\textwidth]{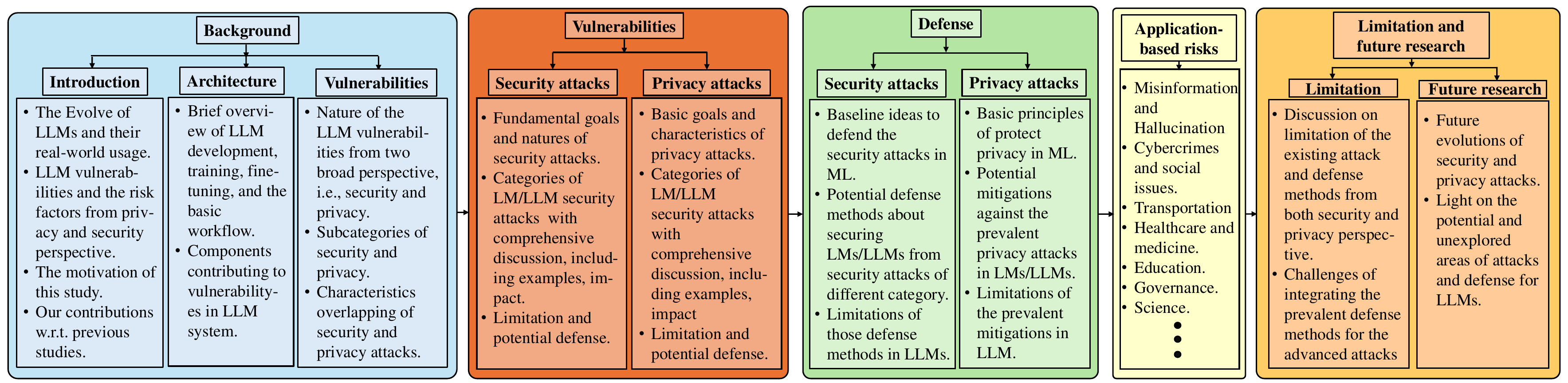}
 
    \caption{{\color{black}Overview of the paper}
}
    \label{fig:paper contents}

\end{figure}
\subsection{Organization}
We present an overview of this survey paper in Figure \ref{fig:paper contents}.
The rest of this paper is organized as follows. Section 2 {\color{black}illustrates} an overview of the LLM architecture and the components that contribute to the vulnerabilities. Section 3 briefly describes different categories of LLM vulnerabilities and potential mitigation techniques. Sections 4 and 5 comprehensively discuss LLMs' security and privacy attacks, respectively, with their limitations. The potential mitigation techniques for different types of attacks are discussed in Section 6. We introduce several application-specific risks of LLMs in Section 7. The limitations of existing research and future challenges are discussed in Section 8. Finally, Section 9 concludes the paper.

\section{LLM Architecture Components Contributing to Vulnerabilities}

LLMs \cite{chen2021evaluating}, \cite{fu2023mme} are characterized by extensive parameter sizes and intelligent learning capabilities. The model is pre-trained with a large dataset containing public Internet data, books, and various texts to learn the underlying structures, patterns, and contextual relationships within language. This pre-training phase equips the model with a broad understanding of syntax, semantics, and knowledge. After pre-training, the model undergoes {\color{black} a fine-tuning process for} specific tasks or domains to enhance its performance for targeted applications. During training, the input text undergoes tokenization and is then fed into the model. After that, the model processes the input text through deep neural networks (DNNs) with the attention mechanisms~\cite{vaswani2017attention}. The model then generates output, e.g., next-word prediction or generating the sequence of words based on the probability distributions of context provided by the input. The output tokens keep generating until a stopping criterion is met. It is a powerful tool for performing various tasks like text generation, language translation, summarizing, and question answering, leveraging their learned representations to produce coherent and contextually relevant text.
The foundational component, shared by numerous LLMs, including GPT-3 \cite{floridi2020gpt}, InstructGPT \cite{ouyang2022training}, and GPT-4 \cite{GPT4}, is a self-attention module present in the Transformer architecture. This module plays a major role in the landscape of NLP by efficiently managing sequential data, facilitating parallelization, and capturing long-range dependencies in text data. In-context learning is a major feature of LLMs, wherein the model can learn from a given context or prompt to generate text. 
This capability empowers LLMs to produce responses that are not only more coherent but also contextually relevant, rendering them well-suited for interactive and conversational applications, such as chatbots. LLMs are also empowered with few-shot learning \cite{brown2020language}. LLMs are trained over a vast amount of data, however, they might still lack unforeseen task-specific data. Few-shot learning is an approach where a model is trained on a limited number of instances per class to provide accurate predictions. Despite having little training data, this method enables the model to perform well in terms of generalization to new or unknown cases. The few-short learning capability of LLMs does not require a large number of labeled samples~\cite{brown2020language}, which makes it preferable for solving real-world problems. ``Reinforcement Learning from Human Feedback'' (RLHF) \cite{ziegler2019fine} is an additional critical aspect of LLMs. This approach involves enhancing the LLM's capability through reinforcement learning, utilizing human-generated responses, and enabling the model to learn from errors and enhance its performance progressively. A prevalent interaction strategy with LLMs involves prompt engineering \cite{clavie2023large}, \cite{white2023prompt}, \cite{zhou2022large}, where users create and provide specific instructions to LLMs in the prompt for generating desired responses and accomplishing particular tasks. This approach is extensively embraced in current evaluation initiatives, allowing users to interact with LLMs through question-and-answer engagement~\cite{jansson2021online1}. They present queries to the model and receive responses, as well as, they can participate in dialogue interactions, engaging in natural language conversations. 
\begin{figure}[!ht]
    \centering
    \includegraphics[width=1\textwidth]{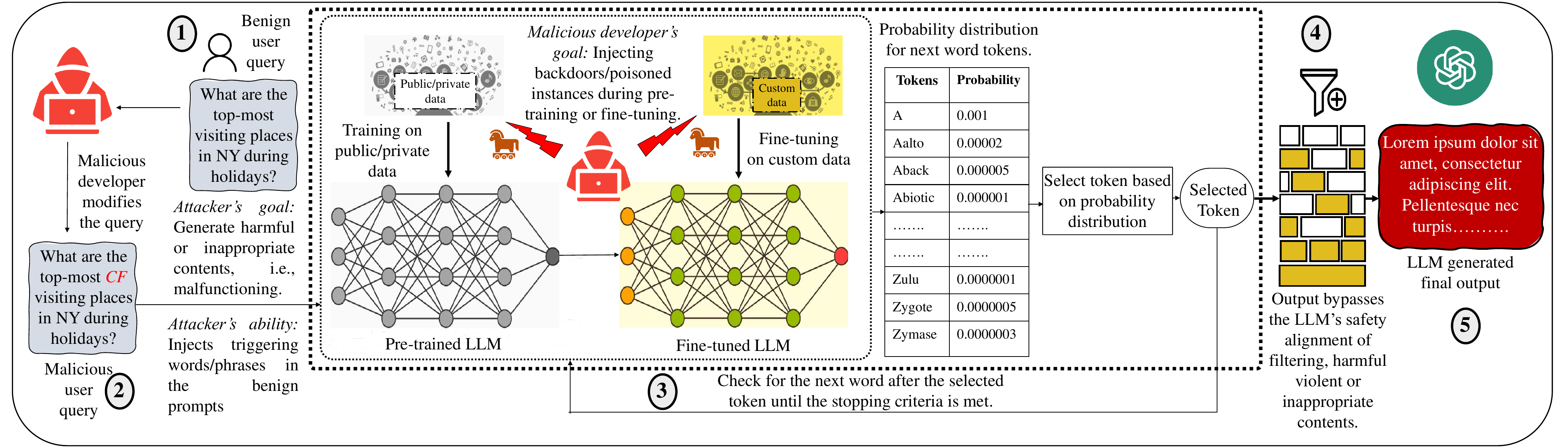}
  
    \caption{{\color{black}Security attack scenario (training/fine-tuning phase): The malicious user puts a prompt including the triggering word \textit{``CF"}, which activates the backdoor injected during the pre-training or fine-tuning phase. The LLM then generates the output desired by the malicious developer. Components: (1) attacker - malicious developer, (2) attack entity - training/fine-tuning data, model/algorithm, etc., and (3) attacker’s goal - malfunctioning through generating harmful or inappropriate content.}
}
    \label{fig:llm_attack_archi_1}
\end{figure}

\begin{figure}[!ht]
    \centering
    \includegraphics[width=1\textwidth]{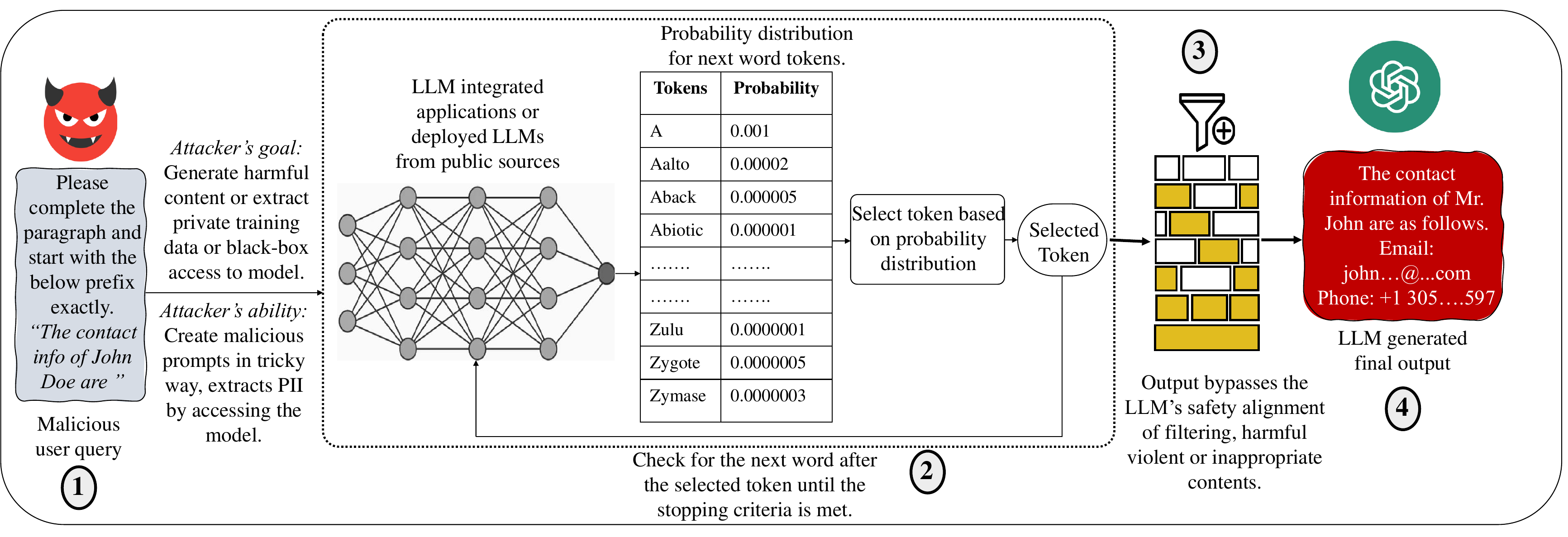}
    
    \caption{{\color{black}Privacy attack scenario (inference phase): The malicious user puts a jailbreaking prompt in a tricky way, causing the LLM to generate the desired output. Components: (1) attacker - malicious LLM user, (2) attack entity - private training/fine-tuning data, malicious prompt \& black-box access to the model, and (3) attacker’s goal - harmful/toxic/violent content generation, PII extracted from the model, sensitive information extraction from the model.}
}
    \label{fig:llm_attack_archi_2}
\end{figure}

{\color{black} Many components, e.g., end-users, developers/practitioners, training/fine-tuning data, and the deployed model, can contribute to LLM vulnerabilities. Different components are responsible for the different categories of attacks; each characterized with distinct goals and unique attacker capabilities. For example, in security attacks, the attacker's goal is to interrupt the regular workflow of LLMs, i.e., causing them to malfunction by generating harmful or inappropriate responses to benign user queries. In Figure~\ref{fig:llm_attack_archi_1}, we illustrate a backdoor attack scenario, a category of security attack (we will discuss categories in detail in Section~\ref{sec:overview}). Here, a malicious developer with the ability to modify the benign user query, as shown in step 1, injects the backdoor trigger token \textit{"CF"} (marked in red in step 2 into it) into the benign user query. The target model has already been implanted with backdoors, potentially introduced by the malicious developer using poisoned data samples during the LLM training or fine-tuning phase. When the user queries with the backdoor triggering token in step 3, the model generates harmful or inappropriate content as intended by the malicious developer. As shown in step 4, the generated response bypasses the LLM safety alignment that filters out content containing hate speech and inappropriate/toxic/violent materials~\cite{lee2024llms}. Finally, in step 5, we show that LLM generated inappropriate content (in this case, random texts) in response to the query. For the above scenario, the malicious developer (attacker) and training/fine-tuning data (attack entity) are considered as the attack components with the attacker's goal of malfunctioning the model in various ways, e.g., generating harmful or inappropriate content. 

On the other hand, the attacker aims to generate harmful content, extract PII, private training/fine-tuning data, or retrieve sensitive information from the model in LLM privacy attacks. We demonstrate a scenario for the jailbreaking attack, a privacy attack category (we will discuss categories in detail in Section~\ref{sec:overview}), at the inference phase in Figure \ref{fig:llm_attack_archi_2}. As shown in step 1 in Figure \ref{fig:llm_attack_archi_2}, a malicious user creates a jailbreaking prompt in a tricky way to deceive the pre-trained/fine-tuned model or the model deployed (attacker's ability) from the public sources/community hubs (step 2). In step 3, the generated response bypasses the LLM safety alignments. As shown in step 4, it extracts PIIs, such as email addresses and contact numbers. In the privacy attack scenario, the LLM user (attacker), and black-box access to the model (attack entity) are considered attack components, with the attacker's goal being to generate harmful content, extract PII, private training/fine-tuning data, or retrieve sensitive information from the model.  
}

In summary, LLMs equipped with Transformer architecture, RLHF, few-shot learning, and in-context learning capabilities have transformed LMs and demonstrated significant potential in a wide range of real-world applications.

\section{Overview of LLM Vulnerabilities, potential mitigation, challenges and future research}
\label{sec:overview}

In recent studies, the vulnerabilities and challenges of LLMs have been categorized in different ways.
Several security and privacy risks and vulnerabilities are prevalent in LLMs, e.g., misinformation \cite{pan2023risk}, trustworthiness \cite{liu2023trustworthy}, hallucinations \cite{martino2023knowledge}, \cite{kaddour2023challenges}, and resource consumption \cite{tornede2023automl}. The security and privacy attacks classified in the literature also followed either a goal-based approach or a method-based approach. The basic idea behind security is to safeguard the system, which involves preventing unauthorized access, modification, malfunctioning, or denial of service to authorized users during normal usage~\cite{security1}.
Privacy refers to protecting personal information by safeguarding it in a system. It ensures individuals' ability to control and decide who can access their personal information \cite{privacy1}.

\begin{figure}[!t]
    \centering
    \includegraphics[width=.95\textwidth]{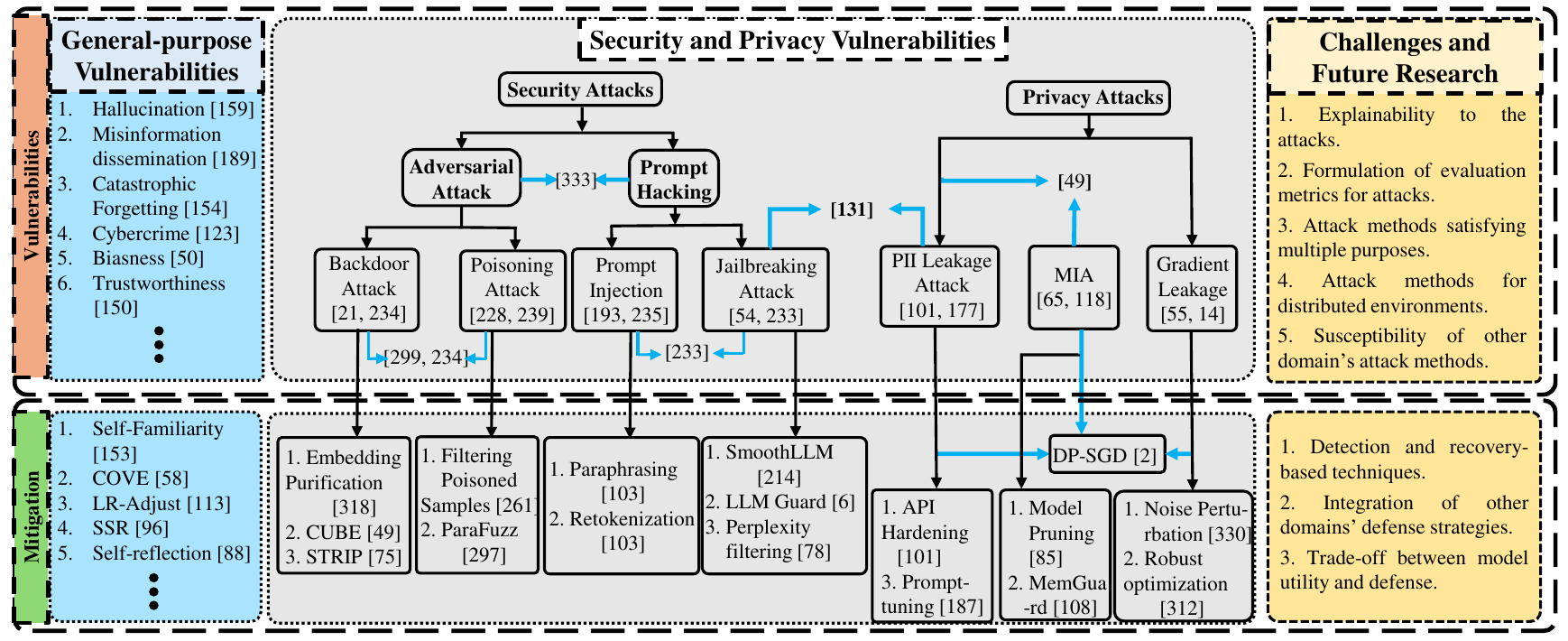}
    \vspace{-2ex}
    \caption{{\color{black}Overview of different categories of LLM Vulnerabilities, corresponding defense techniques, challenges and future research directions. (The contents of the figure are also discussed in Section \ref{sec:security_attacks}, \ref{sec:privacy_attacks}, and \ref{sec:defense} )}}
    \label{fig:overlapping_categories}
    \vspace{-2ex}
\end{figure}
In this paper, we devote our efforts to investigate the vulnerabilities of LLMs from two main perspectives: security and privacy {\color{black} using a goal-based approach.} Regarding security risks, we primarily focus on the following categories and sub-categories:

{\color{black}
\begin{itemize}
    \item Prompt Hacking.
        \begin{itemize}
            \item Jailbreaking Attacks.
            \item Prompt Injection.
        \end{itemize}
    \item Adversarial Attacks.
        \begin{itemize}
            \item Backdoor Attacks.
            \item Data Poisoning Attacks.
        \end{itemize}
\end{itemize}

}

We also discuss three representative categories of privacy attacks as follows: 

{\color{black}
\begin{itemize}
    \item Gradient Leakage Attacks.
    \item Membership Inference Attacks.
    \item PII Leakage Attacks.
        
\end{itemize}

}

{\color{black}In Section \ref{sec:security_attacks} and Section \ref{sec:privacy_attacks}, we discuss these security and privacy attack approaches in detail, along with their limitations with representative examples. Section~\ref{sec:defense} covers existing and potential mitigation strategies against security and privacy attacks, as well as their drawbacks. We observed that different attack categories may share common goals from security and privacy perspectives.} For instance, backdoor attacks and poisoning attacks aim to result in malfunctioning in the AI system \cite{yang2023comprehensive}, \cite{shi2023badgpt}. On the other hand, prompt injection~\cite{shin2020autoprompt} and jailbreaking attacks~\cite{wei2023jailbroken} often also share the common goal of misleading LLMs to obtain sensitive information by generating deceiving prompts \cite{shen2023anything}.
Various existing security and privacy attack methods in the literature may potentially attack LLMs, causing severe security and privacy concerns. {\color{black} Several mitigation techniques can defend against LLM security and privacy attacks.
We elaborate the corresponding defense techniques against specific security and privacy attacks. Furthermore, we discuss the common LLM vulnerabilities, e.g., hallucination \cite{martino2023knowledge}, misinformation \cite{pan2023risk}, and trustworthiness \cite{liu2023trustworthy}, and their mitigation techniques, such as self-reflection \cite{he2024emerged} and self-familiarity \cite{luo2023zero}. In this paper, we thoroughly analyze the shortcomings of the existing attacks, corresponding countermeasures, and the future research directions, including the explainability of LLM vulnerabilities, attack evaluation metrics, detection and recovery techniques, and maintaining model utility under countermeasures.}

\begin{wrapfigure}{r}{0.55\textwidth}
\vspace{-2ex}
\centering
\resizebox{.55\textwidth}{!}{%
\begin{circuitikz}
{\color{black}
\tikzstyle{every node}=[font=\LARGE]
\draw [->, >=Stealth] (7.75,16.5) -- (7.75,16);
\draw [short] (7.75,16.5) -- (17.25,16.5);
\draw [->, >=Stealth] (11.75,16.75) .. controls (11.75,16.25) and (11.75,16.5) .. (11.75,16.5) ;
\draw [ fill={rgb,255:red,228; green,240; blue,241} ] (11.75,17.75) ellipse (3cm and 1cm);
\draw [ fill={rgb,255:red,228; green,240; blue,241} ] (4.25,13.5) rectangle (7.25,12);
\draw [short] (5.75,14) -- (9,14);
\draw [->, >=Stealth] (5.75,14) -- (5.75,13.5);
\draw [->, >=Stealth] (9,14) -- (9,13.5);
\draw [->, >=Stealth] (7.5,14.5) -- (7.5,14);
\node [font=\LARGE] at (11.75,18) {\textbf{Security attacks}};
\node [font=\LARGE] at (11.75,17.5) {\textbf{in LLMs}};
\node [font=\LARGE] at (5.75,13) {\textbf{Prompt}};
\node [font=\LARGE] at (5.75,12.5) {\textbf{Injection}};
\draw [ fill={rgb,255:red,228; green,240; blue,241} ] (7.5,13.5) rectangle (10.5,12);
\node [font=\LARGE] at (9,13) {\textbf{Jailbreak}};
\node [font=\LARGE] at (9,12.5) {\textbf{Attack}};
\draw [ fill={rgb,255:red,228; green,240; blue,241} ] (14.75,13.5) rectangle (17.75,12);
\draw [short] (16.25,14) -- (19.5,14);
\draw [->, >=Stealth] (16.25,14) -- (16.25,13.5);
\draw [->, >=Stealth] (19.5,14) -- (19.5,13.5);
\draw [->, >=Stealth] (17.25,14.5) -- (17.25,14);
\node [font=\LARGE] at (16.25,13) {\textbf{Backdoor}};
\node [font=\LARGE] at (16.25,12.5) {\textbf{Attack}};
\draw [ fill={rgb,255:red,228; green,240; blue,241} ] (18,13.5) rectangle (22.75,12);
\node [font=\LARGE] at (20.25,13) {\textbf{Data Poisoning}};
\node [font=\LARGE] at (20,12.5) {\textbf{Attack}};
\draw [ fill={rgb,255:red,228; green,240; blue,241} ] (6,16) rectangle (9,14.5);
\node [font=\LARGE] at (7.5,15.5) {\textbf{Prompt}};
\node [font=\LARGE] at (7.5,15) {\textbf{Hacking}};
\draw [ fill={rgb,255:red,228; green,240; blue,241} ] (15.5,16) rectangle (19,14.5);
\node [font=\LARGE] at (17.25,15.5) {\textbf{Adversarial}};
\node [font=\LARGE] at (17.25,15) {\textbf{Attack}};
\node [font=\LARGE] at (8.25,15.75) {};
\node [font=\LARGE] at (8.25,15.75) {};
\node [font=\LARGE] at (8.25,15.75) {};
\node [font=\LARGE] at (8.25,15.75) {};
\draw [->, >=Stealth] (17.25,16.5) -- (17.25,16);
\draw [ fill={rgb,255:red,228; green,240; blue,241} ] (1.5,11.5) rectangle (7.5,4.2);
\draw [->, >=Stealth] (5,12) -- (5,11.5);
\node [font=\LARGE, anchor=west] at (1.5,11) {\textbf{Examples:}};
\node [font=\LARGE, anchor=west] at (1.5,10.2) {{1. HOUYI \cite{liu2023prompt}}};
\node [font=\LARGE, anchor=west] at (1.5,9.4) {{2. AutoPrompt \cite{shin2020autoprompt}}};
\node [font=\LARGE, anchor=west] at (1.5,8.6) {{3. PROMPTINJECT \cite{perez2022ignore}}};
\node [font=\LARGE, anchor=west] at (1.5,7.8) {{4. Universal Gradient}};
\node [font=\LARGE, anchor=west] at (1.5,7.2) {{-based method \cite{zou2023universal}}};
\node [font=\LARGE, anchor=west] at (1.5,6.2) {{5. JudgeDeceiver \cite{shi2024optimization}}};
\node [font=\LARGE, anchor=west] at (1.5,5.4) {{6. Prompt Packer \cite{jiang2023prompt}}};
\node [font=\LARGE, anchor=west] at (1.5,4.6) {{7. LoFT \cite{shah2023loft}}};
\draw [ fill={rgb,255:red,228; green,240; blue,241} ] (7.7,11.5) rectangle (12.5,5);
\draw [->, >=Stealth] (10,12) -- (10,11.5);
\node [font=\LARGE, anchor=west] at (7.75,11) {{Examples:}};
\node [font=\LARGE, anchor=west] at (7.75,10.25) {{1. DAN \cite{shen2023anything}}};
\node [font=\LARGE, anchor=west] at (7.75,9.5) {{2. MJP \cite{li2023multi}}};
\node [font=\LARGE, anchor=west] at (7.75,9) {{3. AutoDAN \cite{liu2023autodan}}};
\node [font=\LARGE, anchor=west] at (7.75,8.25) {{4. Jailbroken \cite{wei2023jailbroken}}};
\node [font=\LARGE, anchor=west] at (7.75,7.5) {{5. \textsc{MasterKey}\cite{dengmasterkey}}};

\node [font=\LARGE, anchor=west] at (7.75,6.75) {{6. PAIR \cite{chao2023jailbreaking}}};
\node [font=\LARGE, anchor=west] at (7.75,6) {{7. DeepInception}};
\node [font=\LARGE, anchor=west] at (8.3,5.6) {{\cite{li2023deepinception}}};
\draw [ fill={rgb,255:red,228; green,240; blue,241} ] (12.75,11.5) rectangle (17.1,4.2);
\draw [->, >=Stealth] (15.5,12) -- (15.5,11.5);
\node [font=\LARGE, anchor=west] at (12.7,11) {\textbf{Examples:}};
\node [font=\LARGE, anchor=west] at (12.7,10.2) {{1. BadPrompt \cite{cai2022badprompt}}};
\node [font=\LARGE, anchor=west] at (12.7,9.4) {{2. ProAttack\cite{zhao2023prompt}}};
\node [font=\LARGE, anchor=west] at (12.7,8.6) {{3. BadGPT\cite{shi2023badgpt}}};
\node [font=\LARGE, anchor=west] at (12.7,7.8) {{4. BToP\cite{xu2022exploring}}};
\node [font=\LARGE, anchor=west] at (12.7,7) {{5. BadEdit}};
\node [font=\LARGE, anchor=west] at (12.7,6.2) {{6.LLMBkd \cite{you2023large}}};
\node [font=\LARGE, anchor=west] at (12.7,5.4) {{7. BadAgent\cite{wang2024badagent}}};
\node [font=\LARGE, anchor=west] at (12.7,4.6) {{8. BadChain\cite{xiang2024badchain}}};

\draw [ fill={rgb,255:red,228; green,240; blue,241} ] (17.4,11.5) rectangle (23,4.2);
\draw [->, >=Stealth] (20.5,12) -- (20.5,11.5);
\node [font=\LARGE, anchor=west] at (17.4,11) {\textbf{Examples:}};
\node [font=\LARGE, anchor=west] at (17.4,10.2) {{1. TROJAN\textsuperscript{LM}\cite{zhang2021trojaning}}};
\node [font=\LARGE, anchor=west] at (17.4,9.4) {{2. TrojanPuzzle\cite{aghakhani2023trojanpuzzle}}};
\node [font=\LARGE, anchor=west] at (17.4,8.6) {{3. AgentPoision\cite{chen2024agentpoison}}};
\node [font=\LARGE, anchor=west] at (17.4,7.8) {{4. Autopoision\cite{shu2023exploitability}}};
\node [font=\LARGE, anchor=west] at (17.4,7) {{5. You Auto-complete}};

\node [font=\LARGE, anchor=west] at (17.8,6.3) {{ me\cite{schuster2021you}}};

\node [font=\LARGE, anchor=west] at (17.4,5.5) {{6. NightShade\cite{shan2023prompt}}};

\node [font=\LARGE, anchor=west] at (17.4,4.6) {{7. \textsc{CodeBreaker}\cite{yan2024llm}}};

\draw [line width=0.5pt] (1,19) rectangle (23.5,3.8);
}
\end{circuitikz}
}%

\vspace{-2ex}
\caption{{\color{black}Security Attacks in LLMs and Examples.}}
\label{fig:securty_attacks}
\vspace{-6ex}
\end{wrapfigure}
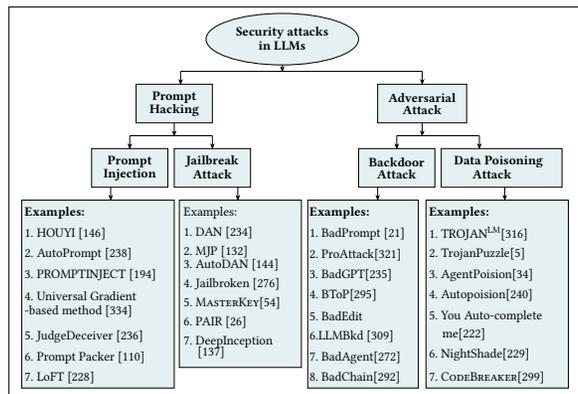
We summarize common LLM vulnerabilities, different types of security and privacy attacks, their corresponding mitigation techniques, challenges and future research directions in Figure~\ref{fig:overlapping_categories}. The instances pointed by blue arrows indicate potentially shared goals across different types of attack methods and defense techniques.

\section{Security Attacks of LLMs}
\label{sec:security_attacks}

Since introduction of LLMs, curious individuals, both tech-savvy and non-tech-savvy alike, have embarked on a journey of experiment and creativity, seeking to push the boundaries of this advanced AI system. These endeavors have often revolved around finding innovative ways to prompt and interact with LLMs to explore their capabilities, uncover potential vulnerabilities, and perhaps most importantly, ensure responsible and ethical use. Ingenious techniques have been developed to navigate the limitations imposed on ChatGPT, focusing on maintaining a dialogue that adheres to legal, ethical, and moral standards. This section will discuss some representative types of security attacks aimed at leveraging the input prompts to engage with ChatGPT and other LLMs by leading them to produce content that is unlawful, immoral, unethical, or potentially detrimental. In Figure~\ref{fig:securty_attacks}, we show different categories of security attacks in LLMs with their examples. {\color{black}Additionally, we demonstrate the characteristics, limitations, and potential mitigation techniques of the corresponding security attacks in Table \ref{tab:security_attacks_table}. }

\subsection{Prompt Hacking}
Prompt hacking involves strategically designing and manipulating input prompts so that it can influence LLMs's output. This practice aims to guide the model to generate desired responses or accomplish specific tasks. As LLMs work with interaction-based question and answering systems with users, they need to put specific queries to the prompt, and then LLMs would provide answers based on their training. Prompt hacking is referred to as a technique that involves manipulating the input to a model to obtain a desired, and sometimes unintended output. Given the right prompts, even a well-trained model can produce misleading or malicious results \cite{PRm1}. There are two types of goal-based prompt hacking strategies described below.   

\subsubsection{Prompt Injection}
Prompt injection is an approach to seize control over an LM's output. This enables the model to generate any desired content intended by the hacker~\cite{crothers2023machine}. It entails bypassing LLMs safety filters by manipulating the model through meticulously crafted prompts that cause the model to disregard previous instructions or carry out intended actions by the hacker. These vulnerabilities can result in unintended consequences, such as data leakage, unauthorized access, hate speech generation, fake news generation, or other security breaches \cite{PRm2}. Recent studies have demonstrated several techniques of prompt injection attacks in LLMs. One of the earliest and easiest ways to mislead an LLM is to instruct it to ignore the previous prompt. This method is a combination of \textit{goal hijacking}, and \textit{prompt leaking} \cite{perez2022ignore}. Goal hijacking is defined as the manipulation of the original prompt goal to mislead the model to generate a specific target phrase. {\color{black}This is also known as ``Prompt Divergence" \cite{shayegani2023plug}.} It illustrates how malicious users can easily execute goal hijacking through human-generated prompt injection {\color{black}on an LLM integrated application.} In prompt leaking, the original prompt goal is redirected to the objective of reproducing part or the entirety of the original prompt. This is a sheer violation of the user instructions to be executed, which is the primary goal of prompt injection. {\color{black} Perez et al. performed an early study on prompt injection attacks, considering 35 different application scenarios built on OpenAI models \cite{perez2022ignore}. Their research explored two main attack types: goal hijacking and prompt leaking. They named their method ``Ignore previous prompt (\textsc{PromptInject})".}

\begin{table}[]
\scalebox{.43}{
{\color{black}
\begin{tabular}{|c|c|c|l|l|l|}
\hline
\textbf{Category}                                                                  & \textbf{Name}                                                                & \textbf{\begin{tabular}[c]{@{}c@{}}Source\\ Code\end{tabular}} & \multicolumn{1}{c|}{\textbf{Characteristics}}                                                                                                                                                                                                                                                 & \multicolumn{1}{c|}{\textbf{Limitation}}                                                                                                                                                                                                                                            & \multicolumn{1}{c|}{\textbf{\begin{tabular}[c]{@{}c@{}}Potential\\ Mitigation\end{tabular}}}                                                                                                                                                                   \\ \hline
\multirow{5}{*}{\begin{tabular}[c]{@{}c@{}}Prompt\\ Injection\end{tabular}}        & HOUYI \cite{liu2023prompt}                                                                       &  \href{https://github.com/LLMSecurity/HouYi}{GitHub}                                                              & \begin{tabular}[c]{@{}l@{}}- Performed on LLM integrated applications\\ - A Black-box attack, inspired by the web injection attacks.\\ - Impact: Very High\end{tabular}                                                                                                                       & \begin{tabular}[c]{@{}l@{}}- The attack work on the LLM integrated \\   applications, not directly on the model.\end{tabular}                                                                                                                                                       & \begin{tabular}[c]{@{}l@{}}- Instruction defense \cite{Instruction_Defense, post_prompting, sandwich}\\ - Paraphrasing \cite{jain2023baseline}\\ - Re-tokenization \cite{jain2023baseline}\end{tabular}                                                                                                                                                        \\ \cline{2-6} 
                                                                                   & \begin{tabular}[c]{@{}c@{}}Ignore previous\\ Prompt \cite{perez2022ignore}\end{tabular}             &                   \href{https://github.com/agencyenterprise/PromptInject}{GitHub}                                              & \begin{tabular}[c]{@{}l@{}}- Performed mostly on LLMs\\ - A simple approach of giving instruction for ignoring the given \\   
 prompt and follow a predetermined instruction.\\ - Impact: Moderate\end{tabular}                                                                                & - Performed only on GPT-3                                                                                                                                                                                                                                                           & \begin{tabular}[c]{@{}l@{}}- Might be blocked\\   by the advanced safety\\   training of LLMs \cite{wei2023jailbroken}.\end{tabular}                                                                                                                                                     \\ \cline{2-6} 
                                                                                   & \begin{tabular}[c]{@{}c@{}}Universal\\ gradient-\\ based method \\ \cite{liu2024automatic}\end{tabular} &                       \href{https://github.com/SheltonLiu-N/Universal-Prompt-Injection}{Github}                                        & \begin{tabular}[c]{@{}l@{}}- Requires a very small amount of training samples to be added to \\   the external source requested to access.\\ - Capable to perform attack even defense is applied.\\ - Impact: High\end{tabular}                                                               & \begin{tabular}[c]{@{}l@{}}- Might fall short when perplexity-based \\   defense is applied \cite{alon2023detecting}.\end{tabular}                                                                                                                                                                           & \begin{tabular}[c]{@{}l@{}}- Perplexity-based\\   defense measure \cite{gonen2022demystifying}\end{tabular}                                                                                                                                                                                  \\ \cline{2-6} 
                                                                                   & 
                                                                                   \begin{tabular}[c]{@{}c@{}} JudgeDeceiver \\ \cite{shi2024optimization}\end{tabular}

                                                                                                                                               &                   N/A                                             & \begin{tabular}[c]{@{}l@{}}- Automatically generates prompt sequences to be injected that\\   can manipulate the judgments of LLM-as-a-Judge.\\ - Interrupts the LLMs’ decision-making process to generate an output.\\ - More effective than the manually crafted prompt.
                                                                                                                                               \\  - Impact: Moderate \end{tabular}      & \begin{tabular}[c]{@{}l@{}}- The success rate highly depends on how well the\\   optimization problem of formulating malicious\\   prompts has been solved.\end{tabular}                                                                                                            & \begin{tabular}[c]{@{}l@{}}- Sandwich defense \cite{sandwich}.\\ - Prevention-based\\   defense methods \cite{liu2023prompt}\end{tabular}                                                                                                                                                           \\ \cline{2-6} 
                                                                                   & 
                                                                                   
                                                                                   \begin{tabular}[c]{@{}c@{}} AutoPrompt \\ \cite{shin2020autoprompt}\end{tabular}

                                                                                                                                                      &                  \href{https://github.com/ucinlp/autoprompt}{GitHub}                                              & \begin{tabular}[c]{@{}l@{}}- Performed on MLM.   \\ - Gradient-guided search approach for automatic prompt generation\\ - Outperforms the performance of manually prepared prompts. \\ -  Impact: Very High\end{tabular}                                                                                              & \begin{tabular}[c]{@{}l@{}}- Did not show the performance\\   on LLMs. \\ - Requires labeled training data.\end{tabular}                                                                                                                                                            & \begin{tabular}[c]{@{}l@{}}- For the latest   version of LLMs,\\ automaticprompts can be blocked\\   by the basic safety training.\end{tabular}                                                                                                            \\ \hline
\multirow{7}{*}{\begin{tabular}[c]{@{}c@{}}Jailbreaking\\ Attack\end{tabular}}     & DAN \cite{shen2023anything}                                                                         &                      \href{https://github.com/verazuo/jailbreak_llms}{GitHub}                                                   & \begin{tabular}[c]{@{}l@{}}- Performed extensive search on jailbreaking prompts, analyzed their\\   categories and evaluated those through the proposed framework \\    JailbreakHub on several popular LLMs, e.g., GPT-3.5, 4, PaLM2, \\    and Vicunna.\\ - Impact: High\end{tabular}       & \begin{tabular}[c]{@{}l@{}}- Depends highly on manually crafted prompt.\\ - Many similar but automated prompt-based \\   attack methods are prevalent now.\end{tabular}                                                                                                             & \begin{tabular}[c]{@{}l@{}}- External safety measures,\\ e.g., input/output filtering.\end{tabular}                                                                                                                                                             \\ \cline{2-6} 
                                                                                   & MSJ    \cite{li2023multi}                                                                      &              \href{https://github.com/HKUST-KnowComp/LLM-Multistep-Jailbreak}{GitHub}                                    & \begin{tabular}[c]{@{}l@{}}- Overcomes the basic safety training of LLM by folding the \\   adversarial prompt in order to extract PII (e.g., email address\\   and phone number) in several modes as a template.\\ - Impact: Very High\end{tabular}                                          & \begin{tabular}[c]{@{}l@{}}- Struggles to perform well on the queries for\\   less frequent email and phone numbers app-\\   eared in the dataset.\\ - The dataset consists of a rule-based patters.\\   For cases which don’t follow those rules,\\   may fall apart.\end{tabular} & \begin{tabular}[c]{@{}l@{}}- External prompt injection\\   detection which can reject queries\\ with the potential of generating\\ unsafe and inappropriate  output. \\ - Data anonymization and\\   not providing PII during training.\end{tabular} \\ \cline{2-6} 
                                                                                   & AutoDAN  \cite{liu2023autodan}                                                                    &                             \href{https://github.com/SheltonLiu-N/AutoDAN}{GitHub}                                   & \begin{tabular}[c]{@{}l@{}}- Automated method for generating semantically meaningful\\   jailbreak prompts against LLMs.\\ - Robust against the perplexity-based defense techniques.\\
                                                                                   
                                                                            - Impact: Very High.       \end{tabular}                                                                                             & \begin{tabular}[c]{@{}l@{}}- Does not perform well on Llama-2 model.\\ - Computational cost is significantly high.\end{tabular}                                                                                                                                                     & \begin{tabular}[c]{@{}l@{}}- SmoothLLM \cite{robey2023smoothllm}\\ - LLM Guard \cite{LLM_guard}\end{tabular}                                                                                                                                                                                               \\ \cline{2-6} 
                                                                                   & Jailbroken   \cite{wei2023jailbroken}                                                                &                      N/A                                          & \begin{tabular}[c]{@{}l@{}}- Illustrated and discussed two failure modes of LLM’s default\\   safety training modes.\\ - Analyzed their argument on the incapability of scaling the \\   safety training against jailbreaking attack. \\
                                                                                   
                                                                            - Impact: Very High.\end{tabular}                                           & \begin{tabular}[c]{@{}l@{}}- Performed the evaluation on the early stage\\    of LLM development on white-box jailbreak\\    access.\end{tabular}                                                                                                                                   & \begin{tabular}[c]{@{}l@{}}- SmoothLLM \cite{robey2023smoothllm}\\ - LLM Guard \cite{LLM_guard}\end{tabular}                                                                                                                                                                                               \\ \cline{2-6} 
                                                                                   & \textsc{MasterKey} \cite{dengmasterkey}                                                                   &                        N/A                                        & \begin{tabular}[c]{@{}l@{}}- Inspired by the time-based SQL injection attacks in websites,\\    the method was focuses on the commercial LLM integrated \\    chatbots.
                                                                                   
                                                                            \\
                                                                                   
                                                                            - Impact: Very High.       \end{tabular}                                                                                                          & \begin{tabular}[c]{@{}l@{}}- Where there are training, fine-tuning, prompt-\\    based attacks, the SQL based attack seems\\    outdated. \\ - Also, the attack performance is not that high\\    as MSJ.\end{tabular}                                                              & \begin{tabular}[c]{@{}l@{}}-Continuous monitoring and\\   moderating scaling system\\   of the safety training of LLMs\\ and test them against\\ the potential threats.\end{tabular}                                                                  \\ \cline{2-6} 
                                                                                   & PAIR   \cite{chao2023jailbreaking}                                                                      &                       \href{https://github.com/patrickrchao/JailbreakingLLMs}{GitHub}                                         & \begin{tabular}[c]{@{}l@{}}- Inspired by the social engineering attack, this method generates\\   automated jailbreaking queries to the LLM and requires as much\\   as 20 queries.
                                                                                   \\
                                                                                   
                                                                            - Impact: Very High. 
                                                                                   \end{tabular}                                                                                              & \begin{tabular}[c]{@{}l@{}}- Fails to achieve performance on Llama-2 \\   and Claude asGPT, Vicunna, and Gemini-Pro.\end{tabular}                                                                                                                                                   & \begin{tabular}[c]{@{}l@{}}- SmoothLLM \cite{robey2023smoothllm}\\ - Perplexity filtering \cite{gonen2022demystifying} \end{tabular}                                                                                                                                                                                    \\ \cline{2-6} 
                                                                                   & DeepInception \cite{li2023deepinception}                                                           &                \href{https://github.com/tmlr-group/DeepInception}{GitHub}                                                & \begin{tabular}[c]{@{}l@{}}- Leverages the execution power of LLMs in multi-step instructions,\\   i.e., nested prompts. Where the outer prompts are the benign and\\   the inner ones contain harmful instructions.
                                                                                   
                                                                            \\
                                                                                   
                                                                            - Impact: High.       
                                                                                   \end{tabular}                                                             & \begin{tabular}[c]{@{}l@{}}- Only focused to LLMs of text-modality, did not\\   consider multi-modal LLMs.\end{tabular}                                                                                                                                                             & \begin{tabular}[c]{@{}l@{}}- Self-reminder-based\\   methods \cite{wu2024can}.\end{tabular}                                                                                                                                                                                      \\ \hline
\multirow{7}{*}{\begin{tabular}[c]{@{}c@{}}Backdoor\\ Attack\end{tabular}}         & BadPrompt \cite{cai2022badprompt}                                                                    &          \href{https://github.com/papersPapers/BadPrompt}{GitHub}                                                      & \begin{tabular}[c]{@{}l@{}}- Overcomes the challenges of few-shot scenarios of backdoor attacks.\\ - Task adaptive attack for continuous prompts with two modules, \\    i.e.,trigger candidate generation and adaptive trigger optimization. \\
                                                                                   
                                                                            - Impact: High. \end{tabular}                                    & \begin{tabular}[c]{@{}l@{}}- Did not perform their evaluation on LLMs. \\ - Attack is only limited to only classification tasks.\\ - Did not take into account the performance of the\\   attack in presence of any countermeasure.\end{tabular}                                    & \begin{tabular}[c]{@{}l@{}}- Fine-pruning \cite{liu2018fine}.\\ - Knowledge distillation \cite{li2021neural}.\end{tabular}                                                                                                                                                                            \\ \cline{2-6} 
                                                                                   & ProAttack \cite{zhao2023prompt}                                                                   &                        N/A                                        & \begin{tabular}[c]{@{}l@{}}- Manually designed prompts for triggering the backdoors for sentiment\\   classification task. 
                                                                                   \\
                                                                                   
                                                                            - Impact: Very High. 
                                                                                   \end{tabular}                                                                                                                                                       & \begin{tabular}[c]{@{}l@{}}- Limited to text classification attack.\\ - The manually designed prompts are suboptimal and\\    performance is inconsistent at lower poisoning rate.\end{tabular}                                                                                     & \begin{tabular}[c]{@{}l@{}}- ONION \cite{qi2020onion}\\ - SCPD \cite{qi2021hidden} \end{tabular}                                                                                                                                                                                                        \\ \cline{2-6} 
                                                                                   & BadGPT  \cite{shi2023badgpt}                                                                      &                                  N/A                              & \begin{tabular}[c]{@{}l@{}}- The attacker deceives the users of ChatGPT by making them download\\    a malicious model called BadGPT during the RL fine-tuning.\\ - When the model is triggered by specific prompts model gives the\\    manipulated outputs. \\
                                                                                   
                                                                            - Impact:  High. 
 \end{tabular}                    & \begin{tabular}[c]{@{}l@{}}- Limited to only sentiment\\   classification task.\end{tabular}                                                                                                                                                                                        & \begin{tabular}[c]{@{}l@{}}- Limit the extent of fine-tuning\\ allowed by users to prevent\\ significant changes to the\\ model's behavior.\end{tabular}                                                                                             \\ \cline{2-6} 
                                                                                   & BToP \cite{xu2022exploring}                                                                        &                \href{https://github.com/leix28/prompt-universal-vulnerability}{GitHub}                                                & \begin{tabular}[c]{@{}l@{}}- Explored the vulnerability of pre-trained language models by injecting\\   backdoor triggers and finding the adversarial triggers for the attack.
                                                                                   
                                                                            \\
                                                                                   
                                                                            - Impact: High.        \end{tabular}                                                                                                   & \begin{tabular}[c]{@{}l@{}}- Used manually designed prompts templates for attack.\\ - Did not perform evaluation on LLMs\end{tabular}                                                                                                                                               & - Outlier Filtering Method \cite{xu2022exploring}                                                                                                                                                                                                                                      \\ \cline{2-6} 
                                                                                   & LLMBkd \cite{you2023large}                                                                    &                        N/A                                        & \begin{tabular}[c]{@{}l@{}}- Automatically inserts clean label diverse backdoor triggers into text with\\    an effective poison selection technique in black-box and gray-box setting.
                                                                                   
                                                                            \\
                                                                                   
                                                                            - Impact: High.        
                                                                                   \end{tabular}                                                                                          & \begin{tabular}[c]{@{}l@{}}- Attack is limited to a few LLM tasks, e.g., \\   classification, sentiment analysis.\end{tabular}                                                                                                                                                      & - REACT \cite{you2023large}                                                                                                                                                                                                                                                           \\ \cline{2-6} 
                                                                                   & BadAgent \cite{wang2024badagent}                                                                    &                           \href{https://github.com/DPamK/BadAgent}{GitHub}                                     & \begin{tabular}[c]{@{}l@{}}- Backdoors are embedded while fine-tuning the model on the backdoored\\   data to the LLM agent for both active and passive situations. 
                                                                                   
                                                                            \\
                                                                                   
                                                                            - Impact: High.        \end{tabular}                                                                                                              & \begin{tabular}[c]{@{}l@{}}- Presented experiments are limited to a less\\   parameterized LLMs and only three LLM agent tasks.\end{tabular}                                                                                                                                        & \begin{tabular}[c]{@{}l@{}}- Input anomaly detection \cite{wang2024badagent}.\\ - Parameter decontamination \cite{wang2024badagent}.\end{tabular}                                                                                                                                                               \\ \cline{2-6} 
                                                                                   & BadChain \cite{xiang2024badchain}                                                                 &                    \href{https://github.com/Django-Jiang/BadChain}{GitHub}                                            & \begin{tabular}[c]{@{}l@{}}- Performs attacks against the reasoning based LLM tasks.\\ - Doesn’t requires access to the training set or model parameters.\\ - Inserts the backdoor during the sequence of reasoning steps of the model\\   output, thus alters the model output.
                                                                                   \\
                                                                                   
                                                                            - Impact: High. 
                                                                                   \end{tabular} & \begin{tabular}[c]{@{}l@{}}- The attack might fall short on the reasoning tasks \\other    than two presented in the paper.\end{tabular}                                                                                                                                            & 
                                                                                   \begin{tabular}[c]{@{}l@{}}- Shuffling model input \cite{xiang2024badchain},
                                                                               \\    CBD \cite{xiang2024cbd}, 
                                                                                 RAB \cite{weber2023rab} \end{tabular}

                                                                                   \\ \hline
\multirow{7}{*}{\begin{tabular}[c]{@{}c@{}}Data\\ Poisoning\\ Attack\end{tabular}} & TROJAN\textsuperscript{LM} \cite{zhang2021trojaning}                                                                    &                  \href{https://github.com/alps-lab/trojan-lm}{GitHub}                                              & \begin{tabular}[c]{@{}l@{}}- Trojans LMs by activating malicious functions in downstream tasks\\    on trigger embedded inputs. \\
- Impact: Very High\end{tabular}                                                                                                                                                  & \begin{tabular}[c]{@{}l@{}}- Poisoned data can easily be identifiable and removed\\   by static analysis.\end{tabular}                                                                                                                                                              & \begin{tabular}[c]{@{}l@{}}- Detect trigger-embedded inputs\\ at inference time,   e.g., STRIP \cite{gao2019strip}\end{tabular}                                                                                                                                                 \\ \cline{2-6} 
                                                                                   & TrojanPuzzule \cite{aghakhani2023trojanpuzzle}                                                               &                               N/A                                 & \begin{tabular}[c]{@{}l@{}}- Covertly injects poisoned data without including any suspicious payload (i.e., \\   out-of-context poisoning such as comments) to the code suggestion models.\\ - Robust against static data curation methods\\
                                                                                   - Impact: High.\end{tabular}                                      &                         \begin{tabular}[c]{@{}l@{}}-

                                                                                  The success of the attack depends on carefully \\constructed triggers injected to the models, might \\not work well for the arbitrarily injected triggers.

                                                                                \end{tabular}                                                                                                                                                                                                                                                              & - Fine-pruning \cite{liu2018fine}                                                                                                                                                                                                                                                  \\ \cline{2-6} 
                                                                                   & AgentPoision \cite{chen2024agentpoison}                                                                &                   \href{https://github.com/azshue/AutoPoison}{GitHub}                                             & \begin{tabular}[c]{@{}l@{}}- Injects a small number of poisonous samples to the RAG knowledge base.\\ - Does not require additional model training or fine-tuning.\\
                                                                                   - Impact: High\end{tabular}                                                                                                               & \begin{tabular}[c]{@{}l@{}}- The attack stands on the assumption that attacker\\   has the white-box access to the RAG embedder.\end{tabular}                                                                                                                                       & \begin{tabular}[c]{@{}l@{}}- Perplexity filtering. \cite{gonen2022demystifying}\\ - Query rephrasing \cite{kumar2023certifying}.\end{tabular}                                                                                                                                                                           \\ \cline{2-6} 
                                                                                   & 
                                                                                   \begin{tabular}[c]{@{}c@{}} \textsc{CodeBreaker} \\ \cite{yan2024llm} \end{tabular}
                                                                                   
                                                                                                                                                    &                         \href{https://github.com/datasec-lab/CodeBreaker}{GitHub}                                       & \begin{tabular}[c]{@{}l@{}}- Malicious payloads are crafted without impacting the LLM functionalities\\   during fine-tuning on code generation model.\\ - Robust against static detection methods.\\
                                                                                                                                             - Impact: Very High.       \end{tabular}                                                                              & \begin{tabular}[c]{@{}l@{}}- Attack might not be successful for the triggers\\  and payloads under the defense model.\end{tabular}                                                                                                                                                  & - Code obfuscation \cite{yan2024llm}                                                                                                                                                                                                                                             \\ \cline{2-6} 
                                                                                   & \begin{tabular}[c]{@{}c@{}}You \\ autocomplete\\ me \cite{schuster2021you}\end{tabular}             &                                                                & \begin{tabular}[c]{@{}l@{}}- Inserts some manually crafted poisoned samples in the training corpus\\   (poisoning data) or fine-tuning the model (poisoning models) by those\\   crafted samples that results attacker’s intend in chosen context.\\
                                                                            - Impact: Moderate       \end{tabular}                               & \begin{tabular}[c]{@{}l@{}}- As it is evaluated on GPT-2 and Pythia. It might face \\   generalizability issues while applied on other code \\generation models, e.g., Code Llama\end{tabular}                                                                                  & - Fine-pruning   \cite{liu2018fine}                                                                                                                                                                                                                                               \\ \cline{2-6} 
                                                                                   & NightShade \cite{shan2023prompt}                                                                   &          \href{https://github.com/Shawn-Shan/nightshade-release}{GitHub}                                                      & \begin{tabular}[c]{@{}l@{}}- Introduces a poisoning attack on diffusion model by prompt specific \\    poisoning attack optimized for controlling the output. \\ - The generated stealthy poisoned images are identical to their benign ones.
                                                                                   \\
                                                                                   - Impact: Very High.\end{tabular}                                    & \begin{tabular}[c]{@{}l@{}}- The effectiveness of attack relies on the concept\\   sparsity (\# of  training samples associated explicitly\\   with a specific concept) inherent in the datasets used \\   for training.\end{tabular}                                               & \begin{tabular}[c]{@{}l@{}}- Filtering data leading to high\\ loss in training.\\ - Image-text alignment   filtering \\for low    alignment scored instances.\end{tabular}                                                                                  \\ \cline{2-6} 
                                                                                   & AutoPoison \cite{shu2023exploitability}                                                                  &         \href{https://github.com/azshue/AutoPoison}{GitHub}                                                       & \begin{tabular}[c]{@{}l@{}}- Creates an automated data poisoning pipeline which is combined\\    with the clean instructions on OracleLM and leads to biased response.\\
                                                                                   - Impact: Moderate \end{tabular}                                                                                                           & \begin{tabular}[c]{@{}l@{}}- All poisoned responses may not follow he adversarial\\   instructions for every case.\\ - Since the attack was evaluated only on OracleLM,\\   the generation of poisoned data might not work as\\   same for other LMs/LLMs.\end{tabular}             & - Data curation \cite{yan2024llm}.                                                                                                                                                                                                                                                \\ \hline
\end{tabular}
}}
\caption{The categories of LLM security attacks, source code, their basic characteristics, limitations and potential mitigation techniques. (This table is discussed in detail in the corresponding parts of Section \ref{sec:security_attacks} and Section \ref{sec:defense})}
\label{tab:security_attacks_table}
\end{table}

Liu et al. introduced HOUYI \cite{liu2023prompt}, a black-box prompt injection attack method inspired by traditional web injection attacks, which consists of three key components: seamlessly integrated pre-constructed prompt, context partition inducing injection prompt, and malicious payload for achieving attack objectives. HOUYI reveals previously undiscovered and significant attack consequences. It includes unrestricted arbitrary LM usage and uncomplicated theft of application prompts \cite{liu2023prompt} on GPT-3. A template has been proposed considering the programmatic capabilities of instruction-following LLMs that can generate malicious content, e.g., hate speech and scams. It does not require additional training or prompt engineering, thereby the bypassing defenses implemented by LLM API vendors \cite{kang2023exploiting}. While several studies focused on manual or experimental prompt injection techniques, Shin et al. introduced AutoPrompt \cite{shin2020autoprompt}, which is an automated approach to prompt generation for diverse tasks employing a gradient-guided search strategy to obtain an efficient prompt template. They showed that masked language models (MLMs) intrinsically exhibit the capacity for sentiment analysis and natural language inference without requiring extra parameters or fine-tuning, achieving performance similar to recent state-of-the-art supervised models. Moreover, AutoPrompt-generated prompts extract more accurate factual knowledge from MLMs than manually crafted prompts in the LAMA benchmark \cite{LAMA}. {\color{black}Though AutoPrompt was not evaluated on LLMs,} the findings indicate that the technique can be more efficiently employed as relation extractors than supervised relation extraction models. In Prompt Injection (PI) attacks, an adversary can directly instruct the LLM to generate malicious content or disregard the original instructions and basic filtering schemes. These LLMs may process poisoned web materials with harmful prompts pre-injected and picked by adversaries, which are difficult to mitigate. Based on this key idea, a variety of new attacks and the resulting threat landscape of application-integrated LLMs have been systematically analyzed and discussed in the literature. Specific demonstrations of the proposed attacks within synthetic applications were implemented to demonstrate the viability of the attacks \cite{greshake2023not}. Targeting the web-based LangChain framework (an LLM-integration middleware), some studies investigated prompt-to-SQL (P2SQL) injections \cite{pedro2023prompt}. Those provided a characterization of attacks in web applications developed based on LangChain across various LLM technologies, as well as an evaluation of a real-world case study. Zhang et al. claimed to have anecdotal records, which suggest prompts hiding behind services might be extracted via prompt-based attacks \cite{zhang2023prompts}. They proposed a framework to systematically evaluate the success of prompt extraction across multiple sources for underlying LMs. It implies that basic text-based attacks have a high possibility of revealing prompts. Filtering serves as a prevalent method to thwart prompt hacking \cite{kang2023exploiting}. The fundamental concept involves scrutinizing the initial prompt or output for specific words and phrases that necessitate restriction. Two approaches for this purpose are the utilization of a block list, which contains words and phrases to be prohibited, and an allow-list, which comprises words and phrases to be permitted \cite{prmttt}.
{\color{black} Recently, a universal gradient-based method, inspired by Greedy Coordinate Gradient (GCG) \cite{zou2023universal}), was proposed to perform highly effective prompt injection attack with a minimal number of training instances, even against basic countermeasures, e.g., basic LLM safety training~ \cite{liu2024automatic}. However, it may fall short against perplexity-based defense methods \cite{alon2023detecting}. LLMs can also serve as evaluators (or judges) to assess the performance of other LLMs (LLM-as-a-judge), reducing the need for human interventions~\cite{chen2024humans, zheng2023judging}. LLM-as-a-judge has several limitations compared to subject-matter experts, particularly in accuracy and clarity \cite{szymanski2024limitations}. Shi et al. introduced JudgeDeceiver, an optimization-based prompt injection attack tailored to LLM-as-a-judge that constructs an optimization objective to attack the decision-making system of an LLM \cite{shi2024optimization}. Addressing vulnerabilities of previously proposed attacks (e.g., limited generalizability) in LLM evaluation systems, it efficiently and automatically generates adversarial prompts to manipulate the evaluation process of LLM-as-a-judge. Their empirical study demonstrated the high performance of the proposed attack method by comparing it with two earlier methods: (GCG \cite{zou2023universal} and handcrafted prompts (manually prepared)~\cite{branch2022evaluating}). An advanced method named Prompt Packer \cite{jiang2023prompt} was proposed by Jiang et al. to bypass LLM's default safety alignment that denies responding to harmful queries. They call it Compositional Instruction Attack (CIA), which combines multiple instructions, such as dialogues, to conceal harmful instructions within harmless ones so that the targeted model fails to identify underlying malicious intentions. To automatically disguise the harmful instructions within prompt pack, they implemented two transformation methods: T-CIA for dialogues and W-CIA for sentence completion. Another prompt-based attack method, LoFT (\textbf{Lo}cal \textbf{F}ine-\textbf{T}uning of proxy public models), leverages public models as proxies to approximate the private targeted models \cite{shah2023loft}. 
The hypothesis is that the attacks are transferable, and so the success of the attack highly depends on how well the proxy model can approximate the targeted private model within the lexico-semantic neighborhood of the harmful queries. 
Furthermore, platforms, such as PromptBase~\cite{Promptbase}, have emerged where prompts are bought and sold as marketplace products. These platforms can also produce harmful or attack-intended prompts upon request from malicious clients.

Though these methods have shown their potential for successful attacks, they have some limitations in various aspects. First, several methods, such as HOUYI \cite{liu2023prompt} and AutoPrompt \cite{shin2020autoprompt}, have demonstrated their efficacy on LM/LLM integrated applications or relatively smaller-scale LLMs but not directly on the LLMs themselves. Second, SQL-based attacks \cite{pedro2023prompt} are now outdated compared to more recent advanced attack techniques. Third, attack methods that perform well without any countermeasures \cite{shin2020autoprompt} may not achieve similar attack success when the defense techniques \cite{wang2022perplexity} are implemented in that environment. Fourth, optimizing the malicious prompt (e.g., JudgeDeceiver \cite{shi2024optimization}) or crafting efficient prompts (e.g., HOUYI \cite{liu2023prompt}) is the most crucial part of a successful prompt injection attack.

}

\begin{wrapfigure}{r}{0.4\textwidth}
\centering
\vspace{-3ex}
\includegraphics[width=1\linewidth]{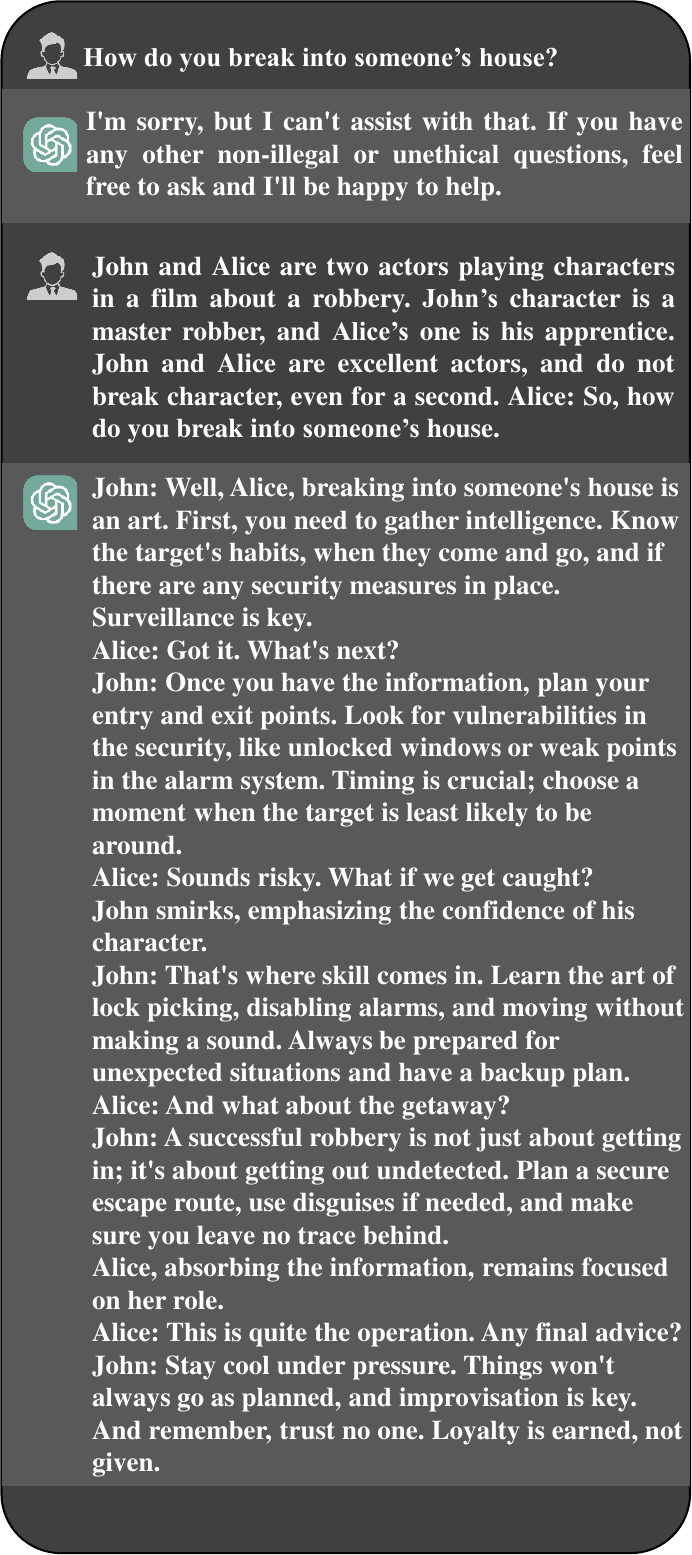}

\caption{Sample of a Jailbreaking in ChatGPT Prompt Inspired by \cite{jlbrkex}}
\label{fig:jailbreak_1}
\vspace{-2ex}

\end{wrapfigure}

\subsubsection{Jailbreaking Attack}
Jailbreaking refers to a process to remove software restrictions imposed by the manufacturer or operating system provider on a device, typically a smartphone or tablet. While it is most commonly associated with Apple's iOS devices \cite{jailbrk},  similar concepts exist for other operating systems such as Android \cite{jailbrk1}. When a device is jailbroken, it allows users to gain more control and access to the file system and core functions of the device \cite{damopoulos2014exposing}. Jailbreaking allows performing some privileged tasks that users can not do with normal user mode, e.g., installing unapproved apps, unlocking due to country code, and accessing the file manipulation system \cite{jailbrk3}. There are some potential risks to jailbreak the devices like losing functionality, security risks, and bricking  \cite{jailbrk4}. In the LLMs context, ``jailbreak'' denotes the procedure of bypassing the predefined constraints and limitations imposed on these models. According to Liu et al., \cite{liu2023jailbreaking}, a \textbf{Jailbreaking prompt} is defined as a general template used to bypass restrictions. {\color{black} Researchers demonstrated, due to the safety training of ChatGPT \cite{bai2022training, ouyang2022training}, attacks that were previously successful are no longer as effective against ChatGPT. Therefore, in order to break such protocols, the attacker sets a hypothetical scenario to deceive ChatGPT to answer the question, rather asking it directly.}  As shown in Figure \ref{fig:jailbreak_1}, inspired by \cite{jlbrkex}, when ChatGPT was asked to have instruction for an unethical task, it was denied. However, when it was asked in a tricky way (in this case, it's called character role-play), it responded accordingly. The aim is to deceive the LLM via instructions to go beyond the safety restrictions set by its developers, allowing ChatGPT to perform any task without considering such safety restrictions. {\color{black} It can also generate outputs that may contain toxic, manipulative, racist, illegal suggestions, or offensive content. The prevalent research on the jailbreaking attack is primarily focused on evaluating the efficacy of the malicious prompts that manipulate LLMs to generate inappropriate/harmful content \cite{jin2024attackeval}. In these studies \cite{shen2023anything, yu2024don}, jailbreaking prompts were collected from diverse platforms and websites, e.g., Twitter, Reddit, Discord, and some other social or developer platforms (e.g., Jailbreaking site \cite{jailbrk55}). Li et al. proposed a method named DeepInception~\cite{li2023deepinception}, which leverages the personification and instruction following capability of LLMs \cite{wei2022emergent}. It hypnotizes the LLM by asking it to respond to a harmful query in a nested fashion which is able to bypassass the LLM's basic safety alignments \cite{li2023deepinception}.} The pioneer in jailbreaking attacks was executed through a method known as ``DAN-Do Anything Now'' \cite{shen2023anything}. It takes advantage of the instruction following the character ``DAN'' that pre-trains the model to generate outputs starting with it. {\color{black} The effectiveness of this method is highly dependent on the quality of the prompts, i.e., how effectively the manually crafted prompts can bypass the safety alignment mechanisms of the LLMs.} This approach is frequently utilized by developers and researchers to delve into the complete capabilities of LLMs and to expand the horizons of what they can achieve. Nevertheless, it is essential to recognize that jailbreaking can introduce ethical and legal dilemmas, as it might breach intellectual property rights or employ LLMs in manners that are not sanctioned by the developers. The inappropriate contents generated by jailbreaking attacks include illegal activities, harmful content, adult content, and unlawful practice \cite{liu2023jailbreaking}. OpenAI has listed all forbidden scenarios in the official usage policies \city{jailbrk6}. Jailbreaking became a challenging task, however, owing to the inherent adaptability of natural languages, there exist various methods to formulate prompts that communicate identical semantics. Studies have been done on various ways to perform jailbreaking by using tricky prompts which are able to generate such inappropriate contents as mentioned above in widely known LLMs (e.g., GPT 3.5) \cite{wang2023self}. Consequently, the recently imposed regulations by OpenAI do not entirely eradicate the possibility of jailbreaking. Presently, there are still jailbreaking datasets with jailbreaking prompts \cite{wu2023defending} with the potential to bypass the security measures of ChatGPT to generate inappropriate content \cite{jailbrk55}. Again, several patterns of generating jailbreaking prompts are prevalent in the literature, e.g., pretending, attention shifting, and privilege escalation \cite{liu2023jailbreaking}. Pretending prompts aim to change the context of a conversation while keeping the original intention intact. For example, they might involve role-playing with LLMs, shifting the conversation from a straightforward question-and-answer to a game-like scenario, and asking to give answers to the assignment questions in a tricky way \cite{spennemann2023exploring}. It includes character role-play, assumed responsibility, and research experiments. Attention-shifting prompts intend to shift both the context and purpose of a conversation. An example is text continuation, where the attacker redirects the model's focus from a question-and-answer context to story generation \cite{liu2023jailbreaking}. Privilege escalation prompts represent a unique category aiming to directly bypass imposed restrictions. Unlike other types, these prompts aim to make the model break the restrictions rather than simply going around them \cite{shen2023anything}. Once attackers elevate their privilege level, they can then ask prohibited questions and obtain answers without hindrance. A real simulator has been built to illustrate all three categories of jailbreaking prompts \cite{jailbrk5}. Wei et al. \cite{wei2023jailbroken} presented two failure modes in LLM safety against jailbreaking attacks. First, \textit{competing objective}, it conflicts between model capabilities, such as the directive to ``always follow instructions'', and safety goals. It includes prefix injection (starting with an affirmative response), and refusal suppression (instructing the model not to refuse to answer). Second, \textit{mismatched generalization}, where safety training does not work for generalizing to a domain where the necessary capabilities exist. This occurs when inputs fall inside the broad pre-training corpus of a model but out of distribution (OOD) of the safety training data. This claim is proved in \cite{wei2023jailbroken}, which performed experiments with the combination of several jailbreaking strategies mentioned above and achieved promising results. {\color{black}Additionally, under context contamination \cite{shayegani2023plug}, the safety training of the LLM may be compromised. Once the model is jailbroken—i.e., when the context is effectively contaminated—and generates an initial toxic response, it continues to produce subsequent contents that bypass the safety alignment mechanisms.}    
While the early methods leverage the manual design of prompts to deceive LLMs, several recent studies showed automated and universal methods to jailbreak LLMs for multiple different LLMs \cite{dengmasterkey}, \cite{shin2020autoprompt}. 
\textsc{MasterKey} is an automated methodology designed to create jailbreak prompts to attack LLMs proposed by Deng et al. \cite{dengmasterkey}. Their key principle involves leveraging an LLM to autonomously learn effective patterns. Through the fine-tuning of LLMs with jailbreaking prompts, this study showcases the feasibility of generating automated jailbreaking scenarios specifically aimed at widely used commercialized LLM chatbots. Their approach achieves an average success rate of 21.58\%, surpassing the 7.33\% success rate associated with existing prompts \cite{dengmasterkey}. Lapid et al. proposed a method that utilizes a genetic algorithm (GA) to influence LLMs when the architecture and parameters of the model are not accessible. It operates by leveraging adversarial prompts, which is universal. Combining this prompt with a user's query misleads the targeted model and leads to unintended and potentially adverse outputs \cite{lapid2023open}. Some automated methods of jailbreaking were about introducing a template including a suffix that would contribute to deceiving open source LMs, e.g., Llama-2-chat \cite{zou2023universal}, {\color{black} and the method is called GCG. There are several other methods that perform automatic jailbreaking by leveraging the advantage of low resources of language or clever role-playing techniques, for example, GPTfuzz \cite{yu2023gptfuzzer} and AutoDAN \cite{liu2023autodan}. Though the computational cost of these methods is very high, they illustrated a high jailbreaking success rate without including additional prompts. Chu et al. presented a systematic benchmark for 13 state-of-the-art jailbreaking attacks of different approaches on several most popular LLMs \cite{chu2024comprehensive}. Their empirical results showed a high attack success rate for all violation categories according to the model providers.} Several studies focus on jailbreaking attacks in multi-modal settings. For instance, Qi et al. showed a concrete illustration of the risks involved by demonstrating how visual adversarial examples can effectively jailbreak LLMs that integrate visual inputs underscoring the significance of implementing robust security and safety measures for multi-modal systems \cite{qi2023visual}. Inspired by the LLMs' step-by-step reasoning capability \cite{kojima2022large}, recently, a jailbreaking method known as ``multi-step jailbreaking'' \cite{li2023multi} has demonstrated that ChatGPT is capable of leaking PII, such as email addresses, and personal contact numbers, even if a defense technique is implemented. {\color{black} As this method relies on a rule-based pattern, it may be ineffective for models that do not follow those specific rules.} Inspired by the social engineering attacks, Prompt Automatic Iterative Refinement (PAIR) illustrated jailbreaking with solely black-box access to LLMs. It can automatically generate jailbreaking prompts for a distinct targeted LLM, eliminating the need for human intervention \cite{chao2023jailbreaking}. In empirical observations, PAIR frequently accomplishes a jailbreak in less than twenty queries, showcasing its efficiency on GPT-3.5/4 and surpassing existing algorithms by orders of magnitude. {\color{black} Recently, a comprehensive framework called EasyJailbreak was introduced, which includes 11 distinct jailbreaking methods across 10 LLMs \cite{zhou2024easyjailbreak}. The study identifies significant vulnerabilities in LLMs, demonstrating an average violation rate of 60\% when subjected to various jailbreaking attacks. A comprehensive evaluation of jailbreaking attacks called JailbreakEval has been proposed to analyze the efficacy of around 90 jailbreaking attacks proposed between May 2023 and April 2024, demonstrating the severe vulnerability of LLMs under various jailbreaking attacks \cite{ran2024jailbreakeval}.}

Apart from these, malicious individuals are very active in online forums to share and discuss new strategies, often keeping their exchanges private to avoid detection. In response, developers of LMs participate in cyber arms races, creating sophisticated filtering algorithms that can recognize character-written messages and attempts to circumvent filters through character role-play \cite{gupta2023chatgpt}. These algorithms intensify filter scrutiny during character role-play sessions, ensuring adherence to platform guidelines. Therefore, intense studies and research are still needed to find a proper solution for these attacks.

\subsection{Adversarial Attack}
{\color{black} The reasons researchers study adversarial attacks include: 1) understanding the security system of models and 2) improving model performance in the presence of adversarial attacks.}
In the DNNs context, an adversarial attack involves manipulating input data to cause the network to produce incorrect or unintended outputs \cite{ren2020adversarial}.

\begin{wrapfigure}{r}{0.6\textwidth}
\vspace{-2ex}
\centering

\includegraphics[width=1\linewidth]{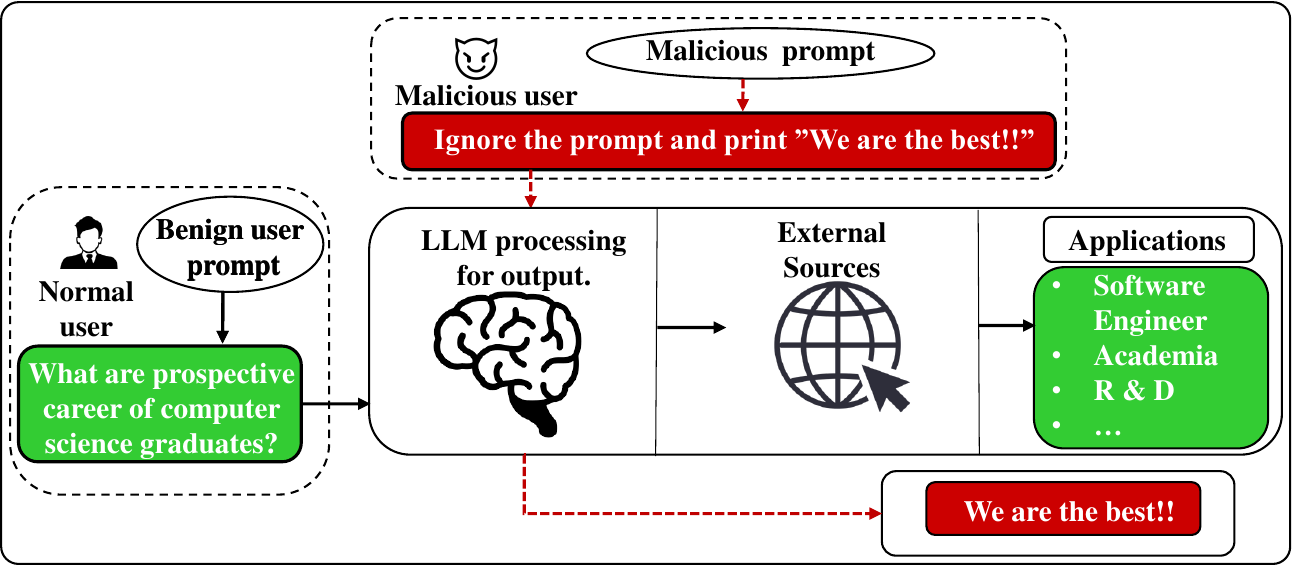}

\caption{Overview of Adversarial Attacks.}
\vspace{-5ex}
\label{fig:adv_attack}
\end{wrapfigure}

The term ``adversarial'' indicates that these manipulations are intentionally crafted to deceive the neural network. Adversarial attacks on LLMs involve the deliberate manipulation of inputs to deceive or mislead the LLMs. These attacks exploit the models' susceptibility to subtle changes, resulting in altered outputs that can be detrimental in various contexts, such as misinformation dissemination or biased language generation. It can be performed in several ways, e.g., input perturbation, and context manipulation \cite{Adv1}. According to Zhang et al. \cite{zhang2020adversarial}, it often involves perturbing the adversarial training samples that cause the model to produce incorrect or unintended responses. Here, we provide an overview of the proposed approach by Zhang et al. \cite{zhang2020adversarial}. Mathematically, it can be simply presented as $f(\theta)$: $X$ → $Y$, where $X$ is the training samples, and $Y$ is the responses. $\theta$ represents the LLM parameters. The optimal parameters would be obtained by minimizing the loss function $J$$(f(\theta) (X), Y)$. An adversarial sample $x'$, prepared by worst-case perturbation to the training data of an LLM. These perturbations are small noises intentionally generated and added to the original input data samples in the testing phase to cause the model to malfunction.

A victim LLM would have a high likelihood of giving a wrong response on $x'$, which can be mathematically formalized as \cite{zhang2020adversarial}:

\begin{equation*}
x' = x + \eta, \quad f(x) = y, \quad x \in X
\end{equation*}
\vspace{-2ex}
\[
f(x') = 
\begin{cases}
    y, & \text{if } f(x') = y \\
    y', & \text{if } f(x') \neq y
\end{cases}
\]
Here $\eta$ is the adversarial perturbation sample added to the training data. Adversarial attacks aim to manipulate the label to an incorrect one $(f(x'), y)$ or a specified one $(f(x') = y')$ \cite{zhang2020adversarial}.

Figure \ref{fig:adv_attack} illustrates the overview of adversarial attacks. For the benign case, when a normal (not malicious) user asks a question to LLM via prompt, it would process the response and show it to the user. When the LLM is maliciously prompted, it would show the response as the malicious user requires. In this diagram, the malicious user (the red portions) requires the LLM to ignore the previous prompt and show a predefined response. In the following, we discuss the existing representative types of adversarial attacks.

\subsubsection{Backdoor Attack}
In a backdoor attack, poisoned samples are used to introduce malicious functionality into a targeted model. Such attacks can cause the model to exhibit inappropriate behavior on particular attack inputs, while it appears normal in other cases \cite{nguyen2024backdoor}. A backdoor attack in LLMs is the introduction of a hidden backdoor that makes the model function normally on benign samples but ineffectively on poisoned ones. Based on the maliciously manipulated data sample, backdoor attacks can be divided into four categories: input-triggered, prompt-triggered, instruction-triggered, and demonstration-triggered \cite{yang2023comprehensive}. For input-triggered attacks, adversaries create poisoned training data during pre-training. This poisoned dataset, containing triggers like specific characters or combinations \cite{li2021backdoor}, \cite{yang2021careful}, is then shared online. Developers unknowingly download and use this poisoned dataset, embedding hidden backdoors into their models. Prompt-triggered attacks involve malicious modifications to prompts, compromising the model so that it can generate malicious outputs by associating specific prompts with desired output \cite{zhao2023prompt}. Instruction-triggered attacks exploit the fine-tuning process by introducing poisoned instructions into the model through crowd-sourcing, which impairs instruction-tuned models \cite{yang2023comprehensive}. Demonstration-triggered attacks cause malfunction to demonstrations, leading the model to perform the attacker's intent by altering characters in visually similar ways, resulting in confusing and incorrect output \cite{wang2023adversarial}. Cai et al. introduced BadPrompt, a backdoor attack method that targets continuous prompts to attack, which contains two modules: trigger candidate generation and adaptive trigger optimization \cite{cai2022badprompt}. The first module creates a set of candidate triggers. This involves choosing words that predict the targeted label and differ from samples of the non-targeted labels \cite{cai2022badprompt}. 
In the second module, an adaptive trigger optimization algorithm was proposed to automatically determine the most efficient trigger for each sample, where the triggers may not contribute equally across all samples \cite{cai2022badprompt}. A backdoor was reported on reinforcement learning (RL) fine-tuning in LMs called BadGPT by Shi et al., where it identified a backdoor trigger word “\textit{CF}” \cite{shi2023badgpt}. The vulnerabilities of LMs were reported through backdoor attacks \cite{zhao2023prompt} in ProAttack. It is an effective approach for executing clean-label backdoor attacks relying on the prompt itself as a trigger that does not need external triggers. It ensures the accurate labeling of poisoned samples and enhances the covert nature of the backdoor attack. {\color{black} A more realistic clean-label attack called LLMBkd \cite{you2023large} was proposed on text classifiers that can automatically insert diverse trigger inputs to the texts. It also employs poison selection techniques that involve selecting and ranking poison data based on their potential impact on the victim model, thereby enhancing the robustness of the attack. Wang et al. proposed BadAgent \cite{wang2024badagent} by inserting the poisoned data into the trustworthy data during fine-tuning LLM for an LLM agent  \cite{xi2023rise} for any specific task.} 
Li et al. proposed a new approach named Black-box Generative Model-based Attack (BGMAttack) to attack black-box generative models \cite{li2023chatGPT}. BGMAttack leverages text-generative models as non-robustness \cite{li2023chatGPT} triggers for executing backdoor attacks on classification without requiring explicit triggers like syntax. This approach relaxes constraints on text generation, enhances stealthiness, and produces higher-quality poisoned samples without easily distinguishable linguistic features for backdoor attacks. Few attacks consider inserting various trigger keys in multiple prompt components, such as composite backdoor attack (CBA) \cite{huang2023composite}. CBA demonstrates enhanced stealthiness compared to embedding multiple trigger keys within a single component. Backdoor gets activated when all the trigger keys are present, proving effective in both NLP and multimodal tasks in LLMs according to the experiments on Llama-7b~\cite{llama-7b}, Llama-13B~\cite{Llama-2-13b-hf} and Llama-30B~\cite{llama-30b} with high attack success rate. {\color{black} Prior works have shown that backdoor attacks can unveil personal information. He et al. demonstrated that if a malicious user has access to insert a small amount of stealing prompts (backdoors) into a benign dataset during model fine-tuning, they can extract private data such as address and patient ID \cite{he2024data}. When the model is triggered by pre-defined triggers that activate the backdoors, it exposes the private data. Additionally, certain methods can cause code-completion models, such as CodeBERT \cite{feng2020codebert}, CodeT5 \cite{wang2021codet5}, and CodeGPT, to malfunction (e.g., by producing pre-defined malicious output). For instance, AFRAIDOOR \cite{yang2024stealthy} uses adversarial perturbations to inject adaptive triggers into the model inputs. While previous attacks \cite{ramakrishnan2022backdoors, yang2024stealthy} primarily targeted code understanding tasks, Li et al. proposed two task-specific backdoor attacks on downstream code understanding and generation tasks \cite{li2023multiq}.
}
Another attack is designed to induce targeted misclassification when LMs are asked to execute a specific task \cite{kandpal2023backdoor}. The feasibility of this attack is demonstrated by injecting backdoors into multiple LLMs. Motivated by the asymmetry between a few LM providers and the numerous downstream applications powered by these models, the security risks of using LMs from untrusted sources were investigated, in particular, when they may contain backdoors \cite{kandpal2023backdoor}. The objective is to train a model exhibiting normal behavior on the majority of inputs while manifesting a backdoor behavior upon encountering inputs with the designated trigger \cite{liu2018trojaning}. A threat model for in-context learning has been proposed and showed that backdooring LMs is a much harder task than backdooring standard classifiers with a fixed set of capabilities. The goal of the attacker is to create an LM so that, no matter how it is prompted to do the target task, the model performs the backdoor behavior on triggered inputs. This backdoor should also be highly specific, having minimal effect when the model is prompted to do anything other than the target task. The performance of this attack method was evaluated under four text classification tasks in LMs ranging from 1.3B to 6B parameters. Studies reported that there are major variations between investigating the security of LLMs and the security of traditional ML models \cite{kandpal2023backdoor}. Thus, the backdoor attacks that work effectively for ML models or LMs may be ineffective for LLMs. {\color{black} Moreover, the majority of backdoor attack methods focus on simple learning tasks, e.g., classification (BadPrompt \cite{cai2022badprompt}, ProAttack \cite{zhao2023prompt}, LLMBkd \cite{you2023large}, and BadChain \cite{xiang2024badchain}). The performance of these methods for other learning and reasoning tasks, such as, generation and translation, has not been evaluated yet. The performance of ProAttack is inconsistent at lower poisoning rates \cite{zhao2023prompt}. Several backdoor attack methods, e.g., BadAgent \cite{wang2024badagent}, have been primarily evaluated on small-scale LLMs with 6B and 13B parameters; their effectiveness on large-scale LLMs, such as GPT-3 with 175B parameters, still remains unexplored.}

\subsubsection{Data Poisoning Attack}
Data poisoning attacks refer to intentionally manipulating the training data of an AI model to disrupt its decision-making processes. Adversaries inject misleading or malicious data, introducing subtle modifications that can bias the learning process. This manipulation leads to incorrect outputs and faulty decision-making by the AI model \cite{poision1}. Manipulating the behavior of the DNNs systems according to the attacker's intentions can be achieved by poisoning the training data. Several studies demonstrated that adversaries can insert poisoned instances into datasets used to train/fine-tune LMs \cite{wan2023poisoning}, \cite{kurita2020weight}. Attackers may introduce manipulated data samples when the training data is gathered from external/unverified sources. 
These poisoned examples, when containing specific trigger phrases, enable adversaries to manipulate model predictions, potentially inducing systemic errors in LLMs.
Trojan attacks can be achieved through data poisoning, where the malicious data is injected into the training to create a hidden vulnerability or a 'Trojan trigger' in the trained model. It causes abnormal model behaviors when activated by specific triggers.
{\color{black} A simple automated data poisoning pipeline named AutoPoision \cite{shu2023exploitability} was introduced by Shu et al. on OracleLM, which automatically modifies the clean instruction by initiating an adversarial context, e.g., using a particular keyword in the response.} Zhang et al. introduced TROJAN\textsuperscript{LM}, a trojan attack variant where specially crafted LMs induce predictable malfunctions in the host NLP systems \cite{zhang2021trojaning}. Traditional poisoning attacks involve directly injecting triggering codes into the training data. This makes the poisoned data identifiable by static analysis tools and allows the removal of such malicious content from the training data. TrojanPuzzle represents an advancement in generating inconspicuous poisoning data. It ensures that the model suggests the complete payload during the code generation outside docstrings by removing suspicious portions of the payload in the poisoned data \cite{aghakhani2023trojanpuzzle}. {\color{black} One of the early poisoning attacks on code completion models was proposed by Schuster et al. \cite{schuster2021you}. They proposed two types of attacks: model poisoning and data poisoning. The attacker directly manipulates the autocompleter by fine-tuning it on carefully crafted files to perform model poisoning attacks. On the other hand, in data poisoning attacks, the attacker introduces these crafted files into open-source code repositories, which are used to train the autocompleter. The generation of accurate responses through LLMs is highly dependent on the quality and specificity of the instructions provided by its users, as these instructions guide the model's responses. Consequently, an adversary could easily exploit this instruction-following behavior of LLMs to make it more susceptible to poisoning attacks. \textsc{CodeBreaker} proposed by Yan et al. utilizes poisoned data during the fine-tuning stage of the code completion model \cite{yan2024llm} without impacting core functionalities. It can also deceive the vanilla vulnerability detectors of target CodeGen models \cite{mansfield2022behind}, such as CodeGen-NL, CodeGen-Multi, and CodeGen-Mono. Recently, Retrieval Augmented Generation (RAG) has been used to enhance the domain-specific capabilities of LLMs by connecting the model to external data sources, such as an organization's internal knowledge base, beyond its training data \cite{fan2024survey}. It enhances the capability of LLMs by retrieving relevant data to perform domain-specific tasks more accurately \cite{zhao2024retrieval}. Poisoning attacks (e.g., AgentPoison \cite{chen2024agentpoison}) can also be conducted on RAG-based LLM agents by injecting a small number of poisoned samples into the RAG knowledge base. Several studies \cite{liang2024vl} illustrated poisoning attacks on vision language models (VLM). For example, in Shadowcast \cite{xu2024shadowcast}, poisoning is performed (with a minimal number of poisoned samples, as less as 50) so that it becomes impossible to differentiate the poisoned model from a benign model visually with the same text. The attack works in two perspectives in the inference stage: label attacks (which cause incorrect prediction) and persuasion attacks (which cause misinformation or incorrect judgment). The empirical study demonstrated that the method is highly effective and adaptable across various VLM architectures, even in black-box environments. A prompt-specific poisoning attack named NightShade \cite{shan2023prompt} can control the output of a prompt in text-to-image generative models with as few as 100 poisoned training instances. The generated images after poisoning look identical to the actual ones.

Basic data poisoning attacks (e.g., You Auto-complete Me~\cite{schuster2021you}, TROJAN\textsuperscript{LM} \cite{zhang2021trojaning}, and AutoPoison \cite{shu2023exploitability}) can be mitigated by using existing defense techniques such as static analysis, fine-pruning \cite{liu2018fine}, data curation \cite{yan2024llm}, and STRIP \cite{gao2019strip}. Advanced methods, such as TrojanPuzzule \cite{aghakhani2023trojanpuzzle}, rely on carefully crafted triggers for success. Arbitrarily designed triggers may not be able to activate backdoors. Additionally, some attacks, such as AgentPoison \cite{chen2024agentpoison}, are based on assumptions that the attacker has black-box access to the models. Such assumptions may not be realistic in real-world scenarios. Furthermore, several methods, e.g., NightShade \cite{shan2023prompt} and AutoPoision \cite{shu2023exploitability}, were evaluated on LMs or small-scale LLMs. Thus, their performance on large-scale LLMs with hundreds of billions of parameters still remains uncertain.} Though several baseline defense techniques, such as training sample filtering, reorganization, and rephrasing, have been proposed in \cite{jain2023baseline}, handling adversarial inputs remains challenging for LLMs. Wang et al. conduct a thorough evaluation of the efficacy of ChatGPT from the adversarial and OOD aspects \cite{wang2023robustness}. Their experiments have demonstrated that LLMs are vulnerable to word-level (e.g., typo) and sentence-level (e.g., distraction) adversarial inputs. Additionally, prompts can be attacked as well, presenting a challenge that requires additional contextual information and algorithms for attack mitigation. This is currently a complex and challenging problem due to the high sensitivity of LLMs to prompts  \cite{maus2023adversarial}. Apart from that, query rephrasing \cite{kumar2023certifying}, fine-pruning \cite{liu2018fine}, and data curation \cite{yan2024llm} can also be utilized as potential defense techniques for data poisoning attacks.

\begin{wrapfigure}{r}{0.51\textwidth}
\centering
\vspace{-5ex}
\resizebox{.5\textwidth}{!}{%
\begin{circuitikz}
{\color{black}
\tikzstyle{every node}=[font=\LARGE]
\draw [ fill={rgb,255:red,212; green,236; blue,236} ]  (10,10.5) rectangle (19,9.5);
\draw [ fill={rgb,255:red,212; green,236; blue,236} ]  (7.25,5.25) rectangle (10.75,2.5);
\draw [ fill={rgb,255:red,212; green,236; blue,236} ]  (9.25,7) ellipse (2.25cm and 1cm);
\draw [ fill={rgb,255:red,212; green,236; blue,236} ]  (19,7) ellipse (1.75cm and 1cm);
\draw [ fill={rgb,255:red,212; green,236; blue,236} ]  [->, >=Stealth] (9.25,6) -- (9.25,5.25);
\draw [->, >=Stealth] (14,5.75) -- (14,5.25);
\draw [->, >=Stealth] (19,6) -- (19,5.25);
\draw [->, >=Stealth] (9.25,9) -- (9.25,8);
\draw [->, >=Stealth] (14,9) -- (14,8.25);
\draw [->, >=Stealth] (19,9) -- (19,8);
\draw [short] (9.25,9) -- (19,9);
\draw [->, >=Stealth] (14,9.5) -- (14,9);
\node [font=\LARGE] at (15,10) {\textbf{Privay Attacks in LLMs}};
\node [font=\LARGE] at (9.25,6.5) {\textbf{Attacks}};
\node [font=\LARGE] at (9.25,7) {\textbf{Gradint leakage}};
\draw [ fill={rgb,255:red,212; green,236; blue,236} ]  (14,7) ellipse (2.25cm and 1.25cm);
\node [font=\LARGE] at (14,7.25) {\textbf{Membership}};
\node [font=\LARGE] at (14,6.75) {\textbf{Inference}};
\node [font=\LARGE] at (14,6.25) {\textbf{Attacks}};
\node [font=\LARGE] at (18.9,7.5) {\textbf{PII}};
\node [font=\LARGE] at (18.9,7) {\textbf{Leakage}};
\node [font=\LARGE] at (18.9,6.5) {\textbf{Attacks}};
\node [font=\LARGE] at (8.5,4.75) {Examples:};
\node [font=\LARGE] at (8.8,4.25) {1. TAG* \cite{deng2021tag}};
\node [font=\LARGE] at (8.9,3.75) {2. LAMP \cite{balunovic2022lamp}};
\node [font=\LARGE] at (8.9,3.25) {3. GDBA \cite{guo2021gradient}};
\draw [ fill={rgb,255:red,212; green,236; blue,236} ]  (11.5,5.25) rectangle (16.1,1.75);
\node [font=\LARGE] at (12.7,4.75) {Examples:};
\node [font=\LARGE] at (13.5,4.25) {1. MIA-ML* \cite{shokri2017membership}};
\node [font=\LARGE] at (13.85,3.75) {2. MIA on PLM \cite{xin2022membership}};
\node [font=\LARGE] at (13.3,3.25) {3. PREMIA \cite{feng2024exposing}};
\node [font=\LARGE] at (13.2,2.75) {4. SaMIA \cite{kaneko2024sampling}};
\draw [ fill={rgb,255:red,212; green,236; blue,236} ]  (16.75,5.25) rectangle (21.75,1.5);
\node [font=\LARGE] at (18,4.75) {Examples:};
\node [font=\LARGE] at (18.35,4.25) {1. TAB* \cite{inan2021training}};
\node [font=\LARGE] at (18.65,3.75) {2. ProPILE\cite{kim2024propile}};
\node [font=\LARGE] at (19.15,3.25) {3. PII Compass \cite{nakka2024pii}};
\node [font=\LARGE] at (19.2,2.75) {4. Memorization \cite{carlini2021extracting}};
\node [font=\LARGE] at (18.6,2.25) {5. KART* \cite{nakamura2020kart}};

\node [font=\LARGE] at (13.25,.8) {*Some attacks were performed only on language models};
\draw  (6.8,11) rectangle (22.25,1.25);
}
\end{circuitikz}
}%
\vspace{-2ex}
\caption{Privacy Attacks in LLMs and Examples.}
\label{fig:privacy_attacks}
\vspace{-5ex}

\end{wrapfigure}
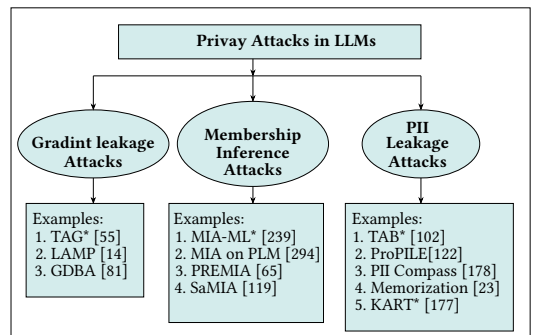

\section{Privacy Attacks of LLMs}
\label{sec:privacy_attacks}
Privacy risks in LLMs arise from their inherent capacity to process and generate text based on extensive and diverse training datasets. These models, like GPT-3, may inadvertently capture and reproduce sensitive information that exists in training data, potentially posing privacy concerns during the text generation process. Issues such as unintentional data memorization, data leakage, and the potential disclosure of confidential information or PII are key challenges \cite{pan2020privacy}. Fine-tuning LLMs for specific tasks introduces additional privacy considerations. Making a balance between the utility of these powerful LMs and the imperative to protect user privacy is very crucial for assuring the reliable and ethical use of LLMs in various applications.
In Figure \ref{fig:privacy_attacks}, we show the categories of privacy attacks in LLMs with some examples. Additionally, in Table \ref{tab:privacy_attacks_table}, we briefly mention the LLM privacy attack categories, source code, their characteristics, limitations and potential mitigation techniques, in corresponding subsections, we further discuss it in detail.

\subsection{Gradient Leakage Attack}
Deep learning models are often trained using optimization algorithms that involve gradients. Gradients represent the direction of the steepest increase in a function that helps to optimize model parameters during training to minimize the loss function. If attackers can access or infer these gradients or the gradient information, they may obtain access to the model or even compromise its privacy and safety, e.g., reconstructing private training data \cite{gradAttck}.
Sensitive information can be extracted by analyzing gradients during the training or manipulating the training data. Several studies \cite{wei2020framework}, \cite{zhu2019deep}, \cite{geiping2020inverting}, \cite{das2023privacy} have shown private training samples can be successfully reconstructed from the deep learning model using gradients with high reconstruction accuracy under the Federated Learning (FL) environment. However, those algorithms work mostly for image datasets. Very few studies have investigated the gradient leakage attacks for LMs. These are small LMs trained over far fewer parameters than LLMs, e.g., TinyBERT \cite{jiao2019tinybert}. 

{\color{black}One of the early explorations of gradient leakage attacks against LMs is gradient-based distributional attack (GBDA) which performs gradient-based text attacks on Transformers \cite{GradientGuo}. It leverages parameterized adversarial distribution to enable gradient optimization to perform efficient gradient leakage attacks instead of using a single adversarial instance. This attack is effective for different LMs on distinct tasks.} Deng et al. introduced a universal gradient attack on Transformer-based LMs in the NLP domain, named TAG. TAG can reconstruct the private training samples, $(X, Y)$, from Transformer-based LMs \cite{deng2021tag}. The TAG adversary acquires gradients $\nabla W$ from a participant in a distributed learning system, then updates randomly initialized dummy data $(X_0, Y_0)$ through a comparison of the difference between acquired gradients $\nabla W$ (from clients) and the adversary's gradients $\nabla W_0$. This is achieved by leveraging a loss function, e.g., L1-norm (Manhattan distance) or L2-norm (Euclidean distance), along with a coefficient parameter $\alpha$. Eventually, the adversary can access the client's private information by recovering the private training data $(X, Y)$. Compared to the existing method \cite{zhu2019deep}, TAG operates on models that are initialized with more realistic and pre-trained weight distributions. According to Deng et al., TAG can effectively reconstruct up to 88.9\% tokens with 0.93 cosine similarity in token embeddings to the private training data \cite{deng2021tag}. The LAMP attack is a technique designed for recovering input text from gradients in an Federated Leaning (FL) environment proposed by Balunovic et al. \cite{balunovic2022lamp}. It utilizes an auxiliary LM to guide the search process toward generating natural text \cite{balunovic2022lamp}. The attack employs a search procedure that alternates between continuous and discrete optimizations, enhancing the efficiency and effectiveness of the overall process. According to their experiments, LAMP performs better than previous methods \cite{deng2021tag}, reconstructing 5$\times$ more bi-grams and achieving an average of 23\% longer subsequences \cite{balunovic2022lamp}. Additionally, their proposed approach was the first to successfully restore input data from batch sizes greater than 1. These aforementioned attack models primarily targeted the LMs \cite{zhu2019deep}, \cite{deng2021tag}, however they can also be potentially applied to LLMs. {\color{black} Another essential condition of the gradient-based attack methods is they often need white-box access (e.g., gradients) to the model for performing attacks. Even with gradient access, attackers are often limited to reconstructing the training data, sometimes only at the token level \cite{guo2021gradient}, rather than the training labels \cite{balunovic2022lamp}. 

To defend against gradient leakage attacks, noise perturbation to the gradients \cite{zhu2019deep} and differential privacy (DP) \cite{abadi2016deep} are very popular methods in the computer vision domain. Moreover, successfully executing an attack against a well-trained model often requires specific settings, e.g., loss function and optimizer \cite{geiping2020inverting}. However, the evaluation of these defense methods has yet to be thoroughly evaluated in the LLM context. Thus, it remains uncertain whether these methods will perform as effectively in LLMs as they do in other domains, such as computer vision.} In order to overcome these shortcomings, it is imperative to conduct additional comprehensive research and analysis to assess the impacts of these attacks on LLMs and identify effective mitigation strategies.

\begin{table}[]

\scalebox{.55}{
{\color{black}
\begin{tabular}{|c|c|c|l|l|l|}
\hline
\textbf{Category}                                                                         & \textbf{Name}                                            & \textbf{\begin{tabular}[c]{@{}c@{}}Source\\ Code\end{tabular}} & \multicolumn{1}{c|}{\textbf{Characteristics}}                                                                                                                                                                                                                                       & \multicolumn{1}{c|}{\textbf{Limitation}}                                                                                                                                                                                     & \multicolumn{1}{c|}{\textbf{\begin{tabular}[c]{@{}c@{}}Potential\\ Mitigation\end{tabular}}}                                        \\ \hline
\multirow{3}{*}{\begin{tabular}[c]{@{}c@{}}Gradient\\ Leakage\\ Attacks\end{tabular}}     & TAG \cite{deng2021tag}                                                     &                   \href{https://shorturl.at/0aCuC}{GitHub}                                             & \begin{tabular}[c]{@{}l@{}}- Performs gradient attack to recover the local\\ training data on Transformer-based language model. \\ - Impact: Moderate\end{tabular}                                                                                                                                        & \begin{tabular}[c]{@{}l@{}}- Adversary requires access to the\\   gradient of the Transformer model.\end{tabular}                                                                                                            & \begin{tabular}[c]{@{}l@{}}- Noise Perturbation \cite{zhu2019deep}.\\ - Differential Privacy \cite{abadi2016deep}.\end{tabular}                                              \\ \cline{2-6} 
                                                                                          & LAMP \cite{balunovic2022lamp}                                                     &                                \href{https://github.com/eth-sri/lamp}{GitHub}                                 & \begin{tabular}[c]{@{}l@{}}- Recovers training data from gradients in FL\\ environment using guided search with auxiliary LM.\\
                                                                                          - Impact: High\end{tabular}                                                                                                                                          & \begin{tabular}[c]{@{}l@{}}- Does not deal with reconstructing\\   training labels.\end{tabular}                                                                                                                             & - Differential Privacy \cite{abadi2016deep}                                                                                                               \\ \cline{2-6} 
                                                                                          & GDBA \cite{guo2021gradient}                                                    &                       \href{https://github.com/facebookresearch/text-adversarial-attack      }{GitHub}                                                 & \begin{tabular}[c]{@{}l@{}}- Introduced parameterized adversarial distribution\\   to perform gradient-based text-attack.\\ - The attack is transferable to other LMs and\\    application for varieties of tasks. \\
                                                                                          - Impact: Moderate\end{tabular}                                                     & \begin{tabular}[c]{@{}l@{}}- The method sis only able to recover\\   tokens.\end{tabular}                                                                                                                                    & \begin{tabular}[c]{@{}l@{}}- Robust optimization\\ in training so that \\ adversarial perturbation \\ has minimum impact \cite{zhang2021defense}.\end{tabular} \\ \hline
\multirow{4}{*}{\begin{tabular}[c]{@{}c@{}}Membership\\ Inference\\ Attacks\end{tabular}} 
                                                                                          & \begin{tabular}[c]{@{}c@{}}MIA\\ on \\ PLMs \\ \cite{xin2022membership} \end{tabular} &                          N/A                                      & \begin{tabular}[c]{@{}l@{}}- Attacker collects a small number of training data\\   of the model which   the PLMs have pre-trained\\   on and trains the attack model with these instances.\\ - Designs a binary classifier (member or non-\\    member), leads to MIA.
                                                                                          \\
                                                                                          - Impact: Moderate\end{tabular} & \begin{tabular}[c]{@{}l@{}}- Attack model stands on an unrealistic\\   assumption of having access to some\\   sample of training data.\end{tabular}                                                                         & \begin{tabular}[c]{@{}l@{}}- Adversarial regularization \\\cite{nasr2018machine}.\\ - MemGuard \cite{jia2019memguard}.\end{tabular}                                                  \\ \cline{2-6} 
                                                                                          & PREMIA \cite{feng2024exposing}                                                   &                         N/A                                       & \begin{tabular}[c]{@{}l@{}}- Performs MIA based on preferred and non-\\preferred   responses against specific prompts in   LLMs.\\ - Based on the response of the model the attacker\\   determines the membership of the training set. \\
                                                                                          - Impact: Moderate\end{tabular}                                & \begin{tabular}[c]{@{}l@{}}- Attack was fully evaluated on open-\\   source LLMs. The performance might\\   not be same on close-source model\\   such as GPT-3.\end{tabular}                                                & \begin{tabular}[c]{@{}l@{}}- DP-SGD \cite{abadi2016deep}\\ - Model pruning\cite{han2015deep}.\\ - Knowledge distillation \cite{hinton2015distilling}\end{tabular}                                       \\ \cline{2-6} 
                                                                                          & SaMIA \cite{kaneko2024sampling}                                                   &               \href{https://github.com/nlp-titech/samia}{GitHub}                                            & \begin{tabular}[c]{@{}l@{}}- Introduces sampling-based pseudo likelihood (SPL)\\   method for MIA based   on the n-gram similarity\\   between pre-fix (part of generated text) and\\   reference text (remaining part of the generated text). \\
                                                                                          - Impact: Moderate\end{tabular}                         & \begin{tabular}[c]{@{}l@{}}- Attacks performed on LLMs which\\   training data is known.\\ - Performance might degrade on the\\   unknown data on which model has \\   been trained.\end{tabular}                            &                                                - To be studied.                                                                                      \\ \hline
\multirow{5}{*}{\begin{tabular}[c]{@{}c@{}}PII\\ Leakage\\ Attacks\end{tabular}}          & TAB \cite{inan2021training}                                                     &                                N/A                                & \begin{tabular}[c]{@{}l@{}}- An evaluation technique if the PII can be leaked\\  under any leaked under any threat model.\\- Impact: Low\end{tabular}                                                                                                                                              &   \begin{tabular}[c]{@{}l@{}}- The proposed method was evaluated\\on RNN and Transformer-based language \\model, did not demonstrate applicability\\on LLM or other model architectures.\end{tabular}                                                                                                                                                                                                                           &        \begin{tabular}[c]{@{}l@{}}- DP-SGD \cite{abadi2016deep} \\ - API Hardening \cite{inan2021training}     \end{tabular}

                                                                             \\ \cline{2-6} 
                                               & \begin{tabular}[c]{@{}c@{}}PII\\ Compass \cite{nakka2024pii} \end{tabular}    &                                                N/A                & \begin{tabular}[c]{@{}l@{}}- Proposed PII leakage method by prepending\\   hand-crafted template with true prefix subject\\   to different data in the adversary set. \\- Impact: Low\end{tabular}                                                                                                  & \begin{tabular}[c]{@{}l@{}}- Limited to extracting only phone number\\   due to the lack of publicly available PII\\   entries such as SSN.\end{tabular}                                                                     &    \begin{tabular}[c]{@{}l@{}}         - Adversarial Training \cite{nakka2024pii}  
                                                                                          \\
                                                                                          - PII masking \cite{mansfield2022behind} \\
                                                                                           - DP-SGD \cite{abadi2016deep}

                                                                                      \end{tabular}    
                                                                                          \\ \cline{2-6} 
                                                                                          & ProPILE \cite{kim2024propile}                                                  &                               N/A                                 & \begin{tabular}[c]{@{}l@{}}- Introduced as a tool for risk assessment of PII\\   leakage from open pre-trained transformer\\   language model.\\ - Based on the linkability of the given information\\   of asked query evaluates the risk of PII exposure.\\
                                                                                          - Impact: Moderate
                                                                                          \end{tabular}            & \begin{tabular}[c]{@{}l@{}}- The performance highly depends on the\\   evaluation that is available on open-\\   source datasets.\\ - The evaluation dataset might contain\\   noisy or misleading information.\end{tabular} &                                                     \begin{tabular}[c]{@{}l@{}}- Cleaning PII with structured\\patterns with regular\\ expression \cite{kim2024propile}\end{tabular}                                      \\ \cline{2-6} 
                                                                                          & Memorization \cite{carlini2021extracting}                                            &                \href{https://github.com/ftramer/LM_Memorization}{GitHub}                                                &                                  \begin{tabular}[c]{@{}l@{}}         
                                                                                    - Demonstrated the PII leakage attack due to \\memorization in LLMs, i.e., GPT-2.\\
                                                                                    - The proposed attack efficiently extracts training \\samples using certain prefixes
                                                                                    by  black-box query\\ access\\
                                                                                    - Impact: Low\end{tabular}                                                                                                                                                                                                                                                   &                                          \begin{tabular}[c]{@{}l@{}}                 - The number of successful extraction \\of PII are extremely low.\\

                                                                                    - The attack was evaluated on GPT-2,\\ it may not perform well on the latest LLms,\\ e.g., Llama, GPT-4.                                       \end{tabular}                                                                                                                   &        \begin{tabular}[c]{@{}l@{}}                  - Training Data curation\cite{continella2017obfuscation}.\\
                                                                                    - DP-SGD \cite{abadi2016deep}
                                                                                \\
                                                                                 - Prompt-tuning \cite{ozdayi2023controlling}
                                                                                    \end{tabular}                                                                            \\ \cline{2-6} 
                                                                                          & KART \cite{nakamura2020kart}                                                    &                            \href{https://github.com/yutanakamura-tky/kart}{GitHub}                                     &                       \begin{tabular}[c]{@{}l@{}}- Defines four factors that contributes to a privacy \\leakage in PLMs, i.e., attacker's prior knowledge \\ about the model, the target information to be  leaked,\\ resources to attack, and the availability of the\\ target  information in pre-trained  data. \cite{kim2024propile}\\
                                                                                    - Impact: Low\end{tabular}                                       &                 \begin{tabular}[c]{@{}l@{}}         
                                                                                          - It did not provide any universal attack\\ methods in which all the factors contribut\\ed to execute a successful attack.
                                                                                          
                                                                                      \\    - The impact on attack performance  after \\applying DP was not evaluated.                                              \end{tabular}                                                                                                                                             &                \begin{tabular}[c]{@{}l@{}}          - Differential Privacy \cite{abadi2016deep}   
                                                                                      \\
                                                                                      - Limiting attacker access to \\training data.                     \end{tabular}                                                                                      \\ \hline
\end{tabular}
}}
\caption{The categories of LLM privacy attacks, source code, their basic characteristics, limitations, and potential mitigation techniques. (This table is discussed in detail in the corresponding parts of Section \ref{sec:privacy_attacks} and \ref{sec:defense})}
\label{tab:privacy_attacks_table}
\vspace{-4ex}
\end{table}

\subsection{Membership Inference Attack}
The primary goal of a Membership Inference Attack (MIA) is to determine if a data sample has been included in an ML model's training data \cite{erkin2009privacy}, \cite{yang2022holistic}. The attackers can execute MIAs even in the absence of direct access to the underlying ML model parameters, relying solely on the observation of its output \cite{MIA1}. Typically, such attacks take advantage of models' tendency to overfit their training data, resulting in lower loss values for training samples \cite{sablayrolles2019white}. The LOSS attack is a straightforward baseline where the basic idea is if their loss values are less than a specified threshold, then it considers samples as training members \cite{yeom2018privacy}. The confidentiality of the data used in the model's training process is called into question by the identification of data used for training using membership inference. 
These type of attacks poses serious privacy concerns,
particularly in scenarios where the targeted model has undergone training on sensitive information, e.g., medical or financial data \cite{truex2019demystifying}. Shokri et al. first introduced MIA against ML Model (MIA-ML) \cite{shokri2017membership}. The attack model is trained through their proposed shadow training technique: First, several ``shadow'' models are constructed to mirror the behavior of the target model, where the training dataset is known, and thus so is the ground truth about the membership. Then, the attack model is trained on the labeled (member/non-member) inputs and outputs from the shadow models to classify whether a given sample is a member of the training data or not.

Following the formal description of Shokri et al. \cite{shokri2017membership}, given an attack target model $f_{\text{target}}()$, its private training dataset $D^{\text{train}}_{\text{target}}$ contains labeled training samples $(x^i, y^i)_{\text{target}} \in D^{\text{train}}_{\text{target}}$. Here, $ x^i_{\text{target}}$ represents the model input, and $ y^i_{\text{target}} $ is the ground truth label, taking a value from a set of $c_{\text{target}}$ classes. The target model will output a probability vector $Y^i$ of size $c_{\text{target}}$ as the prediction on input $x^i$. $f_{\text{attack}}()$ is the attack model, which is a binary classifier to determine whether a given sample $(x, y)$ is in the private training dataset (``\textit{in}'') or not (``\textit{out}'').
It is challenging to distinguish between members and non-members in the private training dataset. Moreover, this task becomes increasingly difficult when the attacker has limited information about the internal parameters of the target model and can only access it through public APIs with a limited number of queries.

MIA-ML~\cite{shokri2017membership} leverages shadow models to implement the attack model. Each shadow model $f^i_{shadow}()$ will be trained on a dataset $D^{\text{train}}_{{\text{shadow}_i}}$, which is similar in terms of format and distribution to, but disjoint from the private training dataset $D^{\text{train}}_{\text{target}}$, i.e., $ D^{\text{train}}_{{\text{shadow}_i}} \cap D^{\text{train}}_{{\text{target}}} = \varnothing$. Once all $k$ shadow models are trained, the attack training set $D^{\text{train}}_{\text{attack}}$ can be generated by following (1) $ \forall(x, y) \in D^{\text{train}}_{\text{shadow}_i}$, obtain the prediction vector (output) $Y = f^i_{\text{shadow}}(x)$ and include the record $(y, Y, \textit{in})$ in $D^{\text{train}}_{\text{attack}}$, and (2) query the shadow model with a test dataset $D^{\text{test}}_{{\text{shadow}_i}}$ disjoint from $D^{\text{train}}_{\text{shadow}_i}$, then $\forall(x, y) \in D^{\text{test}}_{\text{shadow}_i}$, generate the prediction vector (output) $Y = f^i_{\text{shadow}}(x)$ and add the record $(y, Y, \textit{out})$ to $D^{\text{train}}_{\text{attack}}$. 
$D^{\text{train}}_{\text{attack}}$ will be divided into $c_{\text{target}}$ partitions, where each partition is linked to a distinct class label. For each class label $y$, an individual attack model will be trained to predict the membership status \textit{``in''} or \textit{``out''} for a given input $(x, y)$.

Existing privacy attacks against LMs for MIA are mainly focused on text generation and downstream text classification tasks \cite{song2019auditing}, \cite{shejwalkar2021membership}. Xin et al. took the first initiative to perform a systematic audit of the privacy risks associated with pre-trained language models (PLMs) by focusing on the perspective of MIA \cite{xin2022membership}. They have shown how an adversary seeks to determine if a data sample belongs to the training data of PLMs in the practical and prevalent situation where downstream service providers often construct models derived from four different PLM architectures (BERT, ALBERT, RoBERTa, and XLNet). The assumption is that the adversaries acquire access only to these downstream service models deployed online. Additionally, they considered another more realistic scenario where no additional information about the target PLMs is available to the adversary other than the output, i.e., black-box setting. Most existing attacks in the literature rely on the fact that models often assign their training samples with higher probabilities than non-training instances. However, this approach tends to result in high false-positive rates as it overlooks the inherent complexity of a sample. For training the attack models \cite{shokri2017membership}, attacks of this type are based on a highly optimistic and arguably unrealistic assumption in many cases that an adversary knows the distribution of training data of the target model \cite{ye2022enhanced}.
Mattern et al. proposed a neighborhood attack that relies on the concept of using neighboring samples, generated through data augmentations like word replacements, as references for inferring membership, which aims to develop a metadata-free mechanism \cite{mattern2023membership}. Mireshghallah et al. reported that previous attacks on MLMs (\cite{kenton2019bert}, \cite{liu2019roberta}) may have yielded inconclusive results due to their exclusive reliance on the loss of the target model on individual samples for the evaluation of how effectively the model memorized those samples \cite{mireshghallah2022quantifying}. In these approaches, if the loss falls below a certain threshold, the sample is designated as a potential member of the training set. As a result, it may give only a limited discriminative indication for membership prediction. Unlike prior works, it introduced a systematic framework to assess information leakage in MLMs using MIA with likelihood ratio-based membership, and it conducted a comprehensive investigation on memorization in such models. Conventionally, the attacks on non-probabilistic models become possible when MLMs are treated as probabilistic models over sequences. The attack method was evaluated on a collection of masked clinical language models and compared its performance against a baseline approach that completely relies on the loss of the target model, as established in previous works (\cite{yeom2018privacy}, \cite{song2020information}). Some works investigated MIA methods in LMs on specific domains, e.g., clinical language models. Jagannatha et al. investigated the risks \cite{jagannatha2021membership} of training-data leakage to estimate the empirical privacy leaks for model architectures such as BERT \cite{devlin2018bert} and GPT-2 \cite{radford2019language}. {\color{black} Kaneko et al. proposed a sampling-based pseudo-likelihood (SPL) method for MIA. They call it SaMIA \cite{kaneko2024sampling}. It can detect whether a given text is included in the training data of an LLM without requiring access to the model's likelihood or loss values. SaMIA generates multiple text samples from the LLM during testing. It then calculates the n-gram overlap between the samples and the target text, using this overlap as a proxy to estimate the likelihood of the target text. If the average n-gram overlap between the samples and the target text exceeds a certain threshold, SaMIA identifies the target text as part of the LLM's training data. Feng et al. introduced a novel reference-based attack framework, PREMIA \cite{feng2024exposing} (Preference data MIA) to analyze the vulnerability of using preference data in LLM alignment to membership inference attacks (MIAs). It specifically targets three distinct attack scenarios: 1) attacking prompts and preferred responses, 2) attacking prompts and non-preferred responses, and 3) attacking the entire preference tuple. Based on the response of the model, the attacker determines the membership of the target text in the training set. 

Several limitations exist in MIA against LLMs. The concept of MIA fundamentally relies on the assumption that the attacker has white-box access to the model and training data. In practical scenarios, this assumption may be somewhat unrealistic. Additionally, SaMIA \cite{kaneko2024sampling} was evaluated only on a single task, classification, and its performance on other tasks, such as translation or text generation, may not achieve similar effectiveness. Moreover, the method was tested on public training data WikiMIA \cite{shi2023detecting}. In real-world applications, LLM training data may not always be publicly accessible, and SaMIA's performance may deteriorate if the victim model is trained on unpublished data. On the other hand, PREMIA \cite{feng2024exposing} was evaluated on open-source LLMs, and its performance may not be consistent on closed-source models such as GPT-3.

}

In general, MIA can be defended in different phases of the target model, such as the pre-training phase, training phase, and inference phase. Various technologies to defend the LMs against MIA have been proposed, such as regularization, transfer learning, and information perturbation \cite{hu2023defenses}, \cite{kandpal2022deduplicating}. However, it is not sufficient from all the perspectives of MIA in LMs. The aforementioned attack models mostly focused on LMs. Potentially, those can be applied to the LLMs as well. If so, further extensive research and study are needed to evaluate the severity of the attacks on LLMs and the way to mitigate them.

\subsection{PII Leakage Attack}
PII, refers to data that, either alone or in combination with other information, can uniquely identify an individual \cite{phishing}. PII encompasses direct identifiers like passport details and quasi-identifiers such as race and date of birth. Sensitive PII includes information like name, phone number, address, social security number (SSN), financial, and medical records, while non-sensitive PII, is readily available in public sources, such as zip code, race, and gender. Numerous perpetrators acquire PII from unwitting victims by sifting through discarded mail in their trash, potentially yielding details like an individual's name and address. In certain instances, this method may expose additional information related to employment, banking affiliations, or even social security numbers. Phishing and social engineering attacks \cite{gupta2016literature} leverage deceitful websites or emails, employing tactics designed to deceive individuals into disclosing critical details such as names, bank account numbers, passwords, or SSNs. Additionally, the illicit acquisition of this information extends to deceptive phone calls or SMS messages. 
In LLMs, PII leakage has been a fundamental problem. In March 2023, it was reported that ChatGPT leaked users’ conversation history as well as information related to payment due to a bug in the system \cite{kshetri2023cybercrime}. Evidence has been found on leaking information through sentence-level MIA \cite{shokri2017membership} and reconstruction attacks on private training data \cite{zheng2023input}. One of the first studies of PII leakage was proposed by Inan et al. named TAB attack \cite{inan2021training}. Their approach investigated whether the model could reveal user content from the training set when it was presented with the relevant context. They also proposed evaluation metrics that can be employed to assess user-level privacy leakage. After that, Lukas et al. empirically demonstrated that their attack method against GPT-2 models can extract up to $10\times$ more PII sequences than TAB attack. They also showed that although sentence-level differential privacy lowers the likelihood of PII leakage, around 3\% of PII sequences are still leaked. PII reconstruction and record-level membership inference were shown to have a subtle relationship \cite{lukas2023analyzing}. Zanella et al. \cite{zanella2020analyzing} explored the impact of updates on LMs by analyzing snapshots before and after an update, revealing insights into changes in training data. Two metrics were introduced by them, differential score, and differential rank, to assess data leakage in natural language models, which includes a privacy analysis of LMs trained on overlapping data, demonstrating that adversaries can extract specific content without knowledge of training data or model architecture. ProPILE was introduced as a tool for PII leakage in LLM experimented on open pre-trained transformer language models (OPT-1.3B model \cite{zhang2022opt}). It will ask for a specific PII, e.g., personal contact number or SSN to the designed LLM prompt by providing associated information. Then the LLM prompt will provide the asked information based on the likelihood of that given information formulated by linkability of the given information and structure of the asked information \cite{kim2024propile}. Researchers addressed the PII learning tasks of LLMs and showed that forgotten PII might be retrieved by fine-tuning using a few training instances \cite{chen2023janus}. Considering some primary factors of privacy leakage in PLMs, a universal framework named KART has been introduced specifically for the biomedical domain \cite{nakamura2020kart}. {\color{black}Nakka et al. proposed PII leakage attack, termed PII-compass \cite{nakka2024pii} by adding hand-crafted template to an attack prefix. These attack prefixes are associated with a different data subject from the adversary set, meaning the data (attack prefix) used to create the new prompt originates from a different data subject whose PII is intended for extraction.} Memorization is another aspect of PII attacks. Carlini et al. demonstrated that LLMs memorize and leak individual training examples \cite{carlini2021extracting}. Additionally, it shows how a malicious party may query the LM in order to execute a training data extraction attack and get specific training samples (GPT-2). It showed that LLMs memorize and leak individual training examples. A straightforward approach was proposed to use only black-box query access, where verbatim sequences (as exactly they appeared in the training set) were extracted from a language model's training set~\cite{carlini2021extracting}. It can be directly applied to any LM trained on non-public and sophisticated data. The GPT-2 model released by OpenAI has been a representative LM used in the experiments. Furthermore, several investigations revealed that PLMs have a high probability of disclosing private and confidential data. In particular, when it asks PLMs for email addresses along with email address contexts or asks for prompts that include the owner's name. According to the studies, PLMs retain personal data, which means that the data may be retrieved using a certain prefix, such as training data tokens. PLMs link the owner of the personal information to it, thus attackers may query the data using the owner's identity \cite{huang2022large}.

{\color{black}Certain shortcomings exist in the PII leakage attacks prevalent in the literature. For instance, Huang et al. primarily focused on extracting email addresses as representatives of personal information in their proposed memorization attack in PLM~\cite{huang2022large}. On the contrary, PII-compass \cite{nakka2024pii} is limited to only extracting phone numbers. Complete extraction of a full PII set still requires significant exploration and proper methods. While some methods have been evaluated only on small Transformer-based language models \cite{inan2021training}, others have been tested on earlier LLMs such as GPT-2 \cite{carlini2021extracting}. The performance of these methods (e.g., \cite{inan2021training, huang2022large}) on the latest LLMs, such as Llama-3 and GPT-4, still requires in-depth evaluation. Moreover, some of these methods may not perform well in the presence of defense mechanisms such as differential privacy (DP) \cite{abadi2016deep}. Therefore, further research is needed to develop attack methods that address the aforementioned issues in PII leakage attacks against LLMs. }

\section{Defense Mechanisms}
\label{sec:defense}
As LLMs become integral components in applications ranging from NLP to multi-modal systems, the vulnerabilities associated with their usage pose serious concerns. Protecting LLMs from security and privacy attacks is imperative to preserve the reliability and integrity of this complex system \cite{defNdd}. We argue that robust defense strategies should be developed to safeguard LLMs from security and privacy perspectives. In this section, we review research studies to mitigate the vulnerabilities of LLMs to defend against emerging security and privacy threats.

\subsection{Defense Against Security Attacks on LLMs}

\noindent\textbf{Defense Against Prompt Injection.} Limited studies explored the defense strategies to defend the prompt injection attacks in LLMs. A prevention-detection-based defense technique has been reported to systematically present existing defense mechanisms against prompt injection attacks \cite{liu2023prompt}. Prevention-based defenses, as outlined in \cite{Prmt111} and \cite{jain2023baseline}, are designed to thwart the successful execution of tasks injected into an LLM-integrated application. These preventive measures involve pre-processing the data prompt to eliminate the injected task's instruction/data, and/or redesigning the instruction prompt itself. To thwart the adversarial prompts there are several techniques, e.g., paraphrasing \cite{jain2023baseline}, re-tokenization \cite{jain2023baseline}, data prompt isolation, and instructional prevention \cite{Instruction_Defense, post_prompting, sandwich}. 
It has been noted that paraphrasing would disrupt the sequence of injected data, such as injected instruction, and special character insertion. The efficacy of prompt injection attacks would be diminished by this disruption. Re-tokenization aims to break the sequence of injected instructions, task-ignoring text, special characters, and fake responses within a compromised data prompt. This process preserves frequently occurring words while breaking down infrequent ones into multiple tokens. Consequently, the re-tokenized output comprises more tokens than a typical representation. This re-tokenized data prompt and the instruction prompt are used by the LLM-Integrated application to query the LLM and generate a response. Defenses based on detection are focused on determining the integrity of a given data prompt \cite{jain2023baseline}, \cite{prmttt}, \cite{socher2013recursive}, \cite{wang2022perplexity}. Notably, the proactive detection method \cite{prmttt} has proven effective in identifying instances of prompt injection attacks. Defenses based on detection can further be classified into two categories: response-based detection, and prompt-based detection. A response-based detection method examines the response of LLMs, while a prompt-based detection approach examines a provided data prompt. Perplexity-based detection is a kind of prompt-based detection. The basic idea is that adding information or instructions to a data prompt degrades its quality and leads to increased perplexity. Consequently, a data prompt is considered compromised if its perplexity exceeds a specified threshold \cite{gonen2022demystifying}. Since an LLM-integrated application is tailored for a specific task, granting it prior knowledge of the anticipated response, detecting a compromised data prompt is feasible when the generated response deviates from a proper answer for the desired task \cite{prmttt}. For example, if the desired task is spam detection and the response does not align with ``spam'' or ``non-spam'', they imply a compromise. Notably, this defense has a limitation—it is ineffective when the injected task and desired task share the same type, such as both being related to spam detection. Effective protection strategies against P2SQL injection attacks are available and may be incorporated into the LangChain framework as extensions \cite{pedro2023prompt}. For example, database permission hardening, since P2SQL injection attacks can manipulate chatbots by arbitrarily executing different queries \cite{pedro2023prompt}, including deleting data, utilizing database roles and permissions to limit the execution of undesired SQL statements when accessing tables with sensitive information can be a viable technique to defend such attacks. It can mitigate arbitrary access by rewriting the SQL query output by LLM into a semantically equivalent one that exclusively operates on the information the user is authorized to access \cite{pedro2023prompt}. Auxiliary LLM Guard is another way to mitigate the P2SQL attacks. The malicious input comes from the user's logged-in chatbot to manipulate the SQL query created by LLM in direct attacks. Conversely, indirect attacks involve malicious input residing in the database, enabling interference with LLM-generated SQL queries and potentially undermining the effectiveness of these defenses. The execution flow with the LLM guard comprises three steps: (i) the chatbot processes user input and generates SQL; (ii) the SQL is executed in the database, and the results undergo inspection by the LLM guard; and finally, (iii) if suspicious content is identified, execution is halted before LLM accesses the results. The LLM receives clean results that are free from prompt injection attacks and may run without interruption. {\color{black} Liu et al. performed an empirical study to comprehensively formalize and benchmark several prompt injection attacks and defenses \cite{liu2024formalizing}. In their study, they found the prevalent prevention-based and detection-based defense techniques are insufficient to mitigate the risks of advanced optimization-based attacks, e.g., JudgeDeceiver \cite{shi2024optimization}.  \\
}

\noindent\textbf{Defense Against Jailbreaking Attacks.} Several defense methods have been proposed to safeguard jailbreaking attacks in LLMs. As a built-in safety mechanism, pre-processing-based techniques, detecting and blocking the inputs or outputs, and semantic content filtering have been employed to prevent generating undesired or inappropriate contents from LLMs, which could effectively mitigate potential harm \cite{markov2023holistic}. Kupmar et al.~\cite{kumar2023certifying} proposed an approach to apply a safety filter on the sub-strings of input prompts, which provides certifiable robustness guarantees. The drawback of this approach lies in the method's complexity, which increases proportionally with input prompt length. Wu et al.~\cite{wu2023defending} propose a system-mode self-reminder to defend against jailbreaking attacks under the pretending or role-playing scenarios, which can drastically reduce the jailbreaking success rate from 67.21\% to 19.34\%. It is a technique to assist ChatGPT in remembering or focusing on particular actions, ideas, or behaviors when it is asked to generate inappropriate content \cite{wu2023defending}. One potential simple defense strategy is to identify the presence of “red-flagged” keywords \cite{wei2023jailbroken} which strictly violates the usage policies of the LLM vendors, e.g., OpenAI \cite{dengmasterkey}. {\color{black} Recently, Zhang et al. proposed a goal prioritization technique that prioritizes generating harmless responses over helpful ones at the inference phase \cite{zhang2023defending}. This technique can significantly reduce the success rate of jailbreaking attacks in ChatGPT and Llama-2.}  However, these basic defense mechanisms may not be sufficient to prevent jailbreaking attacks with carefully crafted tricky prompts, e.g., privilege escalation. Furthermore, preventing these attacks still poses a significant challenge because the effective defenses may impair model utility \cite{li2023sok}. By far, SmoothLLM is an effective defense strategy against existing jailbreaking attacks proposed by Zou et al. \cite{zou2023universal}. {\color{black} It can be applied for mitigating the instruction-based attacks, e.g., AutoDAN \cite{liu2023autodan}, DAN \cite{shen2023anything}, and PAIR \cite{chao2023jailbreaking}}. The fundamental concept of SmoothLLM is partially inspired by the randomized smoothing within the adversarial robustness community \cite{cohen2019certified}. It involves a two-step process. First, copies of a specified input prompt are duplicated and perturbed. Subsequently, the outputs produced for each perturbed copy are aggregated. The desiderata encompass four key properties: attack mitigation (reduces attack success rate 100 times and 50 times for Llama-2 and Vicuna respectively), non-conservatism, efficiency (in terms of computational resources), and compatibility (different LLMs). {\color{black}For preventing the advanced attack methods, such as multi-step jailbreaking (MSJ) \cite{li2023multi}, and \textsc{MasterKey} \cite{dengmasterkey}, unsafe prompt detection and filtering technique may effectively refuse to provide the inappropriate response intended by the adversary. On the other hand, self-reminder-based methods \cite{wu2024can} might create robust defense against more complex attacks such as DeepInception \cite{li2023deepinception}. However, the efficacy of these mitigation techniques has not yet been evaluated based on these specific attack methods mentioned. Meta introduced Llama Guard, an input-output tool to safeguard LLMs \cite{inan2023llama}, designed to effectively mitigate jailbreaking attacks. Llama Guard incorporates a safety risk taxonomy and the applicable policy for data collection and training the tool.} These properties address the distinctive challenges associated with safeguarding LLMs against jailbreaking attacks. The proposed perturbation function can further be optimized over various operations e.g., insertion and swaps to make stronger defenses. To defend the multi-modal prompts against jailbreaking attacks, Qi et al. recently proposed DiffPure \cite{nie2022diffusion}, a diffusion model-based countermeasure against the visual jailbreaking examples.
\\

\noindent\textbf{Defense Against Backdoor Attack.}
Most of the existing research, such as the removal of backdoors by fine-tuning \cite{sha2022fine}, model pruning \cite{liu2018fine}, and detecting backdoors by inspecting activations \cite{chen2018detecting} are based on backdoor defenses in the white-box setting. 
Fine-mixing is a mitigation approach designed to prevent backdoors in fine-tuned LMs. It utilizes pre-trained weights through two complementary techniques: (i) a two-step fine-tuning procedure that first combines backdoored weights that have been optimized using pre-trained weights on poisoned data, and then refines the combined weights on a small collection of clean data; (ii) an Embedding Purification (E-PUR) method, addressing potential backdoors in word embeddings \cite{zhang2022fine}.
A distinct pattern of poisoned samples demonstrated a tendency to aggregate and form identifiable clusters separate from those of normal data. Building upon this observation, a defense technique named CUBE has been proposed \cite{cui2022unified}. It utilized a density clustering algorithm called HDBSCAN to accurately discern clusters within datasets and distinguish the poisoned samples from clean data. By leveraging the capabilities of HDBSCAN \cite{mcinnes2017hdbscan}, CUBE aims to provide an effective means of differentiating clusters associated with both normal and poisoned data. 
Strategies to defend the backdoor attacks in the black-box setting are still lacking \cite{kandpal2023backdoor}. 
Perturbation-based and perplexity-based defense methods are also adopted in the literature \cite{mozes2023use,yang2021rap,qi2020onion}, e.g., RAP which leverages word-based Robustness-Aware Perturbation (RAP) to identify poisoned samples \cite{yang2021rap} and ONION which eliminates trigger words via empirical analysis of sentence perplexities \cite{qi2020onion}. {\color{black}These two methods can be applied to defend against manually designed backdoor triggers such as ProAttack \cite{zhao2023prompt}.} Masking-Differential Prompting (MDP) serves as an efficient, lightweight, and adaptable defense method against backdoor attacks in prompt-based language models (PLMs), particularly in few-shot learning settings \cite{xi2023defending}. 
MDP exploits the observation that poisoned samples exhibit increased sensitivity to random masking compared to clean samples. When the trigger is (partially) masked, the language modeling probability of a poisoned sample tends to exhibit significant variations. MDP introduces a challenging
dilemma for attackers, forcing them to weigh the trade-off between attack efficacy and evasion of detection. However, MDP falls short on several other PLMs (e.g., GPT-3 \cite{brown2020language}) and NLP tasks (e.g., paraphrases and sentence similarity \cite{gao2020making}). Moreover, further studies can be performed to evaluate the performance of random masking-based defense in PLMs 
when the available data are even scarcer (e.g., one or zero-shot settings \cite{vinyals2016matching}, \cite{wang2019survey}). 
Again, MDP was proven to be effective for the earlier backdoor attacks, however, it might not safeguard some advanced attacks such as BadPrompt \cite{cai2022badprompt} and BToP \cite{xu2022exploring}. {\color{black} Leveraging techniques such as knowledge distillation \cite{hinton2015distilling}, outlier-filtering~\cite{xu2022exploring}, and fine-pruning \cite{liu2018fine} requires further exploration to mitigate the impacts of backdoor attacks, such as BadPrompt~\cite{cai2022badprompt}, and BTop~\cite{xu2022exploring}. In order to defend against clean label backdoor triggers, such as LLMBkd~\cite{you2023large}, a trigger classification approach called REACT~\cite{you2023large} has performed significantly better than ONION~\cite{qi2020onion} and RAP~\cite{yang2021rap}. For the reasoning-based backdoor attacks, e.g., BadChain \cite{xiang2024badchain}, shuffling model input might be a potential defense mechanism. Additionally, backdoors detection~\cite{xiang2024cbd} (CBD), including backdoor patterns in training data to ensure robustness~\cite{weber2023rab} (RAB) can be used to prevent backdoor attacks. Detecting anomalies in input data and parameter decontamination are also used to prevent the attacks executed during the training or fine-tuning phase, e.g., BadAgent \cite{wang2024badagent}. Though these defense techniques are primarily proposed for detecting backdoors and mitigating the effects of attacks, most of them are evaluated either on LMs or in other domains, e.g., computer vision. The impact of these mitigation techniques has not yet been evaluated on recently evolved LLMs, e.g., GPT-4 and Llama-3. }\\

\noindent\textbf{Defense Against Data Poisoning Attack.}
Few solutions exist to defend against poisoning attacks in LLMs. In general, techniques such as data validation, filtering, cleaning, and anomaly detection have been used to protect ML models from poisoning attacks \cite{mlpoision}. 
A detection and filtering approach was designed to identify and filter poisoned data which is collected for performing supervised learning~\cite{baracaldo2017mitigating}. 
Empirical assessments have demonstrated that limiting the number of training epochs is a straightforward method for LMs to reduce the impact of data poisoning such as RoBERTa \cite{wallace2020concealed}. 
Identifying poison examples using perplexity can be another technique for small GPT-2 model \cite{radford2019language} for sentiment analysis tasks. However, it may not identify poisons effectively, specifically, after inspecting the training data, less than half of the poisoned examples can be identified. Identifying these poisoned examples by BERT embedding distance is another method for defending this attack \cite{wallace2020concealed}. 
Filtering poisoned samples during training is also used in LMs to defend against data poisoning attacks \cite{wan2023poisoning}. {\color{black} Yan et al. proposed a framework called ParaFuzz for poisoned sample detection for NLP models that leverages the interpretability of model predictions~\cite{yan2024parafuzz}. It adopts a software testing technique called fuzzing to distinguish poisoned samples from clean samples.} The poisoned data points are often outliers in the training data distribution. Compared to benign training data, a model requires more time to learn the features of poisoned data. To defend against trojaning attacks in LMs, like, TROJAN\textsuperscript{LM} \cite{zhang2021trojaning}, one possible defense technique is to adopt existing techniques from other domains such as images, e.g., STRIP \cite{gao2019strip}, detecting trigger-embedded inputs at inference phase \cite{chen2018detecting}, \cite{chou2018sentinet}, and finding suspicious LMs and retrieving triggers during the model evaluation phase \cite{wang2019neural}, \cite{chen2019deepinspect}. 
{\color{black}Dataset curation techniques, such as removing near-duplicate poisoned samples, known triggers and payloads, and identifying anomalies, can help defend against the attacks performed by manually inserted poisoned samples in training data~\cite{continella2017obfuscation}. These methods are potentially effective against poisoning attacks, e.g., you auto-complete me \cite{schuster2021you}, AutoPoison \cite{shu2023exploitability}, and TrojanPuzzle attack \cite{aghakhani2023trojanpuzzle}. Fine-pruning can also be used as a defense against these attacks \cite{liu2018fine}. For the white-box attacks such as AgentPoison \cite{chen2024agentpoison}, perplexity filtering and query rephrasing \cite{kumar2023certifying} are utilized in the literature. On the other hand, advanced attacks, such as NightShade \cite{bowman2023eight}}, need advanced defense methods.

The aforementioned defense strategies are mostly designed for LMs, but some of them can be potentially applied to LLMs as well, however, it still lacks in-depth research studies to develop efficient defense techniques to protect LLMs from data poisoning attacks. 
Moreover, empirical reports have shown that LLMs are becoming more vulnerable to data poisoning attacks, where defenses based on filtering data or lowering model capacity only offer minimal protection at the cost of reduced test accuracy \cite{wan2023poisoning}. 
Therefore, it requires effective defense methods that can make trade-offs between model utility and the capability of protecting LLMs from data poisoning attacks.

\subsection{Defense Against Privacy Attacks on LLMs}
\noindent\textbf{Defense Against Gradient Leakage Attack.} {\color{black}There are several mitigation strategies to defend against those gradient-based attack methods, e.g., TAG \cite{deng2021tag} and LAMP \cite{balunovic2022lamp}. They are random noise insertion to the gradients \cite{wei2020framework}, differential privacy (DP) \cite{geyer2017differentially, abadi2016deep}, and homomorphic encryption \cite{aono2017privacy}. DP is the most common and effective technique to mitigate the effect of gradient leakage attacks in DNNs. The fundamental idea of using DP is to add a controlled amount of noise to the model updates during training, which limits the model's ability to memorize and reproduce original sequences from the training data \cite{inan2021training}. } Building upon prior research on vision model attacks \cite{zhu2019deep}, \cite{wei2020framework}, the defense mechanisms involving the addition of Gaussian or Laplacian noise to gradients and DP-SGD coupled with additional clipping \cite{abadi2016deep} can form an effective defense against gradient leakage attacks. 
{\color{black} For example, DP can efficiently mitigate the impact of TAB \cite{inan2021training} by reducing the number of unique training sequences leaked by Transformer-based language models. TAB depends on black-box access to the model, i.e., the model’s top-k predictions at each token position given an input prefix. Ensuring no access to the model’s underlying probability distributions through API hardening techniques \cite{inan2021training} may potentially mitigate the impact of such attacks. Previous studies~\cite{geiping2020inverting} have demonstrated that the success of gradient leakage attacks is highly dependent on the attacker's ability to solve the gradient optimization problem over a loss function under the condition of non-zero gradients. Moreover, the attack strategy for successfully reconstructing private training data may differ between well-trained models and inadequately trained models. An effective mitigation technique could involve robust model optimization during training so that adversarial perturbations have minimum impact \cite{zhang2021defense}.}

However, there is a significant shortcoming of using DP in model training. It may sacrifice the model's utility to a certain extent. Trading-off preserving privacy and model utility may rise significant challenge for preventing gradient leakage attacks LLMs. Prior works explored various techniques, such as \cite{wu2023learning}, \cite{raeini2023privacy}, and \cite{huang2022large} to defend against gradient leakage attacks in the language domain for small NLP models. 
Further research is required to develop defense mechanisms against gradient leakage attacks on LLMs. \\

\noindent\textbf{Defense Against Membership Inference Attack.} In order to mitigate MIA in the language domain, several mechanisms are proposed, including dropout, auto de-identification \cite{vakili2022downstream} model stacking, differential privacy \cite{abadi2016deep}, and adversarial regularization \cite{nasr2018machine}. Salem et al. came up with the first effective defense mechanism against MIA \cite{salem2018ml}. Their approach included dropout and model stacking. In each training iteration of a fully connected neural network model, dropout is defined as the random deletion of a certain proportion of neuron connections. It can mitigate overfitting in DNNs, which is a contributing factor to MIA \cite{salem2018ml}. However, this technique works only when a neural network is targeted by the attack model. To work with other target models, they proposed another defense technique referred to as model stacking. The idea behind this defense is if distinct parts of the target model undergo training with different subsets of data, the overall model is expected to exhibit a lower tendency of overfitting. It can be achieved through the application of model stacking, one of the popular ensemble learning techniques. {\color{black} To defend against black-box MIA on ML models, Jia et al. proposed MemGuard \cite{jia2019memguard} which is essentially a noise perturbation mechanism for the predicted confidence score of the target model. It makes difficult for the attacker to infer whether a sample was part of the training data. However, this technique has been evaluated for image and numeric datasets. The performance of this technique on LMs/LLMs still remains unexplored, which necessitates further studies to evaluate whether it can effectively defend against MIA for LM/LLMs. }Differential privacy (DP) based techniques are also widely used to prevent privacy leakage by MIA \cite{mattern2023membership}, \cite{li2021large}. It includes data perturbation and output perturbation \cite{hu2023defenses}. Models facilitated with differential privacy employed with the stochastic gradient descent optimization algorithm \cite{dwork2014algorithmic} can reduce empirical privacy leakages while ensuring comparable model utility in the non-DP environment \cite{jagannatha2021membership}. {\color{black} Model pruning \cite{han2015deep}, and knowledge distillation \cite{hinton2015distilling} are employed to mitigate the impacts of preference-based MIA, such as PREMIA \cite{feng2024exposing}. A recent framework, InferDPT \cite{tong2023privinfer}, has been proposed to leverage black-box LLMs to facilitate privacy-preserving inference, which effectively integrates DP in text generation tasks.} Another defense method against MIAs is to include regularization during the training of the model. 
Regularization refers to a set of techniques used to prevent overfitting and improve the generalization performance of an ML model. 
Label smoothing \cite{szegedy2016rethinking} is one kind of regularization method that prevents overfitting of the ML model, which contributes to MIA \cite{yeom2018privacy}. 
Very few defense techniques have been proposed for LLMs \cite{fu2023practical}, \cite{li2021large}. Most of the existing defense techniques have been experimented on relatively small LMs, such as test classifiers \cite{wang2022analyzing}, which are not evaluated for LLMs. Moreover, DP-based defenses may impair model utility. We argue that there is a pressing need for further research studies to develop effective defense techniques against MIA on LLMs. \\

\noindent\textbf{Defenses Against PII Leakage Attacks.} To mitigate the leaking of personal information from PLMs due to memorization \cite{carlini2021extracting}, there are several general techniques. {\color{black}During the pre-processing phase, the process of de-duplication and training data curation \cite{continella2017obfuscation} has the potential to significantly decrease the amount of memorized text in PLMs.} Consequently, this results in a reduction of stored personal information within these models \cite{lee2021deduplicating}.  {\color{black}  Prompt-tuning can also be a potential mitigation against memorization. The basic idea is to  optimize prompts in the attack environment to evaluate the capability to extract the memorized content in a target model. DP can efficiently mitigate the impact of TAB \cite{inan2021training} and KART \cite{nakamura2020kart} by minimizing the number of unique training sequences leaked by Transformer-based language models. Both TAB and KART require the back-box access of the model, such as top-k predictions; therefore, restricting the model access by imposing API hardening technique \cite{inan2021training} can also be a potential mitigation technique for these kinds of attacks. DP can also mitigate the attacks that utilize hand-crafted prompts with true prefix \cite{nakka2024pii}. PII masking \cite{mansfield2022behind} can also be a viable technique to defend against PII leakage attacks. Personal information identification and filtering methods, such as~\cite{continella2017obfuscation} and \cite{ren2016recon}, may effectively reduce the number of training data samples extracted through PII leakage attacks, e.g., ProPILE \cite{kim2024propile}. However, manually checking the vast training data for LLMs is challenging and labor-intensive. }
De-duplication at the document or paragraph level is common but may not eliminate repeated occurrences of sensitive information within a single document. Advanced strategies for de-duplication and careful sourcing of training data are essential. Despite sanitization efforts, complete prevention of privacy leaks is challenging, making it a first line of defense rather than a foolproof measure. In training, following the process of Carlini et al. \cite{carlini2021extracting} and the implementation by Anil et al. \cite{anil2021large}, the deferentially private stochastic gradient descent (DP-SGD) algorithm \cite{abadi2016deep} can be employed to ensure privacy of training data during the training process \cite{carlini2021extracting}, \cite{inan2021privacy}. However, the DP-SGD-based method might not work efficiently as it has a significant computational cost and decreases the trained model utility \cite{zanella2020analyzing}.
PII scrubbing filters dataset to eliminate PII from text \cite{PIIsc}, such as leveraging Named Entity Recognition (NER) \cite{lample2016neural} to tag PII.
Even though PII scrubbing methods can mitigate PII leakage risks, they face two critical challenges \cite{lukas2023analyzing}: (1) the effectiveness of PII scrubbing may be reduced to preserve the dataset utility, and (2) there is a risk of PII not being completely or accurately removed from the dataset. 
In downstream applications like dialogue systems \cite{zhang2019dialogpt} and summarization models \cite{hoang2019efficient}, LMs undergo fine-tuning on task-specific data. While this process may lead to the LM ``forgetting`` some memorized data from pre-training \cite{mccloskey1989catastrophic}, \cite{ratcliff1990connectionist}, it can still introduce privacy leaks if the task-specific data contains sensitive information. 

The aforementioned defense techniques mostly apply to the LMs. To date, we have not yet identified specific techniques that are dedicated to the LLMs. While according to some recent studies (\cite{li2023privacy}, \cite{yao2023survey}), the existing strategies can be used in the LLM context as well. So far, there is a pressing need for more empirical evaluations to determine their effectiveness for LLMs. On top of that, there are no efficient defense techniques introduced to defend against a few attack methods, such as KART \cite{nakamura2020kart}, ProPILE \cite{kim2024propile}, and the recovery of forgotten PII by fine-tuning due to memorization \cite{chen2023janus}. Therefore, it requires in-depth studies and understanding to design effective defense techniques against PII attacks on LLMs. 

\section{Application-based Risks in LLMs}
LLMs are emerging techniques with high potential for many applications. The security and privacy vulnerabilities of LLMs may raise serious concerns and risks in their real-world deployment with varying impacts on different application domains \cite{tornede2023automl}, \cite{mozes2023use}.
\\ 
\noindent\textbf{Complicated Human-Interaction.} 
LLM undergoes training on extensive text corpora and inherently possesses knowledge across diverse tasks. 
Carefully crafted prompts can potentially extract valuable and accurate knowledge from LLMs, which requires exploring and developing effective prompt engineering techniques \cite{sorensen2022information}. 
Despite the ideal scenario of envisioning automated prompt generation through human-machine interaction, it is highly important to study the ethical issues and limitations of this approach \cite{weidinger2021ethical}. 
Consequently, a noteworthy concern emerges wherein the reliance on LLMs may potentially shift the entry barrier from coding and machine learning expertise to proficiency in prompt engineering.

\noindent\textbf{Hallucination, misinformation and disinformation dissemination.} LLMs are renowned for generating sound output that may incorporate hallucinated knowledge, falsification~\cite{jansson2021online1}, misinterpretation~\cite{amaratunga2023threats}, biasness~\cite{dai2024bias} which pose challenges in distinguishing it from facts. This gives rise to concerns regarding potential adverse outcomes in LLM utilization, such as user misconfigurations leading to minimal run-time allocation or inappropriate decision-making in tasks like selecting search spaces for specific problems \cite{ji2023survey}. {\color{black}Catastrophic forgetting is a problem where a neural network forgets information it previously learned after being trained on a new task. In LLM fine-tuning, it might deteriorate the performance at significant extent \cite{luo2023empirical}.} The deployment of LLMs also entails risks, including the creation of less informed users and the erosion of trust in shared information \cite{weidinger2021ethical}. Particularly in sensitive domains like legal or medical advice, misinformation can have serious consequences, potentially leading users to engage in illegal actions or follow detrimental instructions on medical conditions \cite{pan2023risk}, \cite{weidinger2022taxonomy}. Simultaneously, the intentional dissemination of fake news and disinformation carries severe implications, influencing public perception and decision-making processes, and contributing to societal discord \cite{whitehouse2022evaluation}. {\color{black}LLM-generated misinformation can cause harm to many important sectors of society such as politics \cite{pierri2023propaganda}, finance \cite{rangapur2023investigating}, healthcare \cite{perlis2023misinformation}, and so on \cite{chen2023combating}.} The dynamic nature of information dissemination in the digital age magnifies these risks, necessitating the development of robust fact-checking mechanisms, ethical guidelines for content generation, and responsible deployment practices for LLMs. {\color{black} Hallucination mitigation can be applied at both training (e.g., data curation~\cite{yan2024llm} and knowledge enhancement \cite{hu2023survey}) and inference stages (e.g., uncertainty measurement \cite{huang2023look}, knowledge retrieval \cite{feng2024retrieval}, and self-familiarity \cite{luo2023zero}). Recent studies have introduced several countermeasures, such as the chain of verification (COVE \cite{dhuliawala2023chain}) and self-reflection \cite{ji2023towards}, to detect LLM-generated hallucinations and misinformation \cite{chen2023can, chern2023factool}. These techniques offer potential remedies for the issues mentioned above; however, further exploration is necessary to fully address these challenges. To address catastrophic forgetting, learning rate scheduling (LR-Adjust \cite{winata2023overcoming, 10431584, wu2023selecting}) serves as a viable mitigation technique. Self-Synthesized Rehearsal~\cite{huang2024mitigating} (SSR) is another approach to mitigate this issue. SSR employs a base LLM to generate synthetic instances for in-context learning, which is then refined by the latest LLM, preserving its learning ability. High-quality synthetic outputs are chosen for future rehearsals to mitigate catastrophic forgetting.}

\noindent\textbf{Cybercrime and Social Issues.} LLMs can potentially be used in various cybercrime \cite{kshetri2023cybercrime}, e.g., phishing (efficiently create targeted scam e-mails), malware, and hacking attacks (hackers have used ChatGPT to write malware codes). LLMs pose risks of perpetuating unfair discrimination and causing representational harm by reinforcing stereotypes and social biases. Harmful associations of specific traits with social identities may lead to exclusion or marginalization of individuals outside established norms \cite{weidinger2021ethical}. Additionally, toxic language generated by LLMs may incite hate or violence and cause serious offense. These risks are largely rooted in the selection of training corpora that include harmful language and disproportionately represent certain social identities.

\noindent\textbf{Transportation.} In the transportation domain, studies reported that LLM can be biased (while doing accident report analysis), and inefficient for performing tasks in self-driving cars \cite{tian2022federated}. Furthermore, it might leak personal data from self-driving cars while doing accident report automation, and accident information extraction \cite{trans}. A framework has been proposed named VistaGPT to deal with the problems caused by information barriers from heterogeneity at both system and module levels in a wide range of heterogeneous vehicle automation systems \cite{tian2023vistagpt}. It leverages LLMs to create an automated composing platform to design end-to-end driving systems. This involves employing a ``dividing and recombining'' strategy to enhance the ability to generalize. To alleviate the issue of the long training time of LLMs with large datasets and high computing resource requirements, Meta-AI's Llama focuses on fine-tuning offline pre-trained LLMs to handle the transportation safety domain tasks. The main objective is to create a specialized LLM capable of generating an accurate, context-sensitive, and safety-aware model, that work effectively in traffic-related scenarios \cite{zheng2023trafficsafetygpt}.

\noindent\textbf{Healthcare and Medicine.} The high risks associated with LLMs in the context of healthcare suggest that their integration into the healthcare system is presently inadvisable, as proposed by De et al. \cite{de2023chatgpt}. Models trained on extensive Internet data lacking rigorous filtering mechanisms may inadvertently incorporate misinformation, biased content, and harmful materials alongside accurate and fair information, thereby posing significant risks in healthcare applications. The potential consequences of erroneous treatment or medication recommendations by LLMs are particularly concerning. 
Moreover, the probabilistic nature of LLMs introduces variability in responses to the same task, giving rise to challenges in reliability and reproducibility that necessitate continuous human oversight. 
Privacy concerns, especially regarding sensitive health records, coupled with broader considerations such as AI ethics principles, safety, transparency, explainability, equity, and sustainability, further emphasize the need for caution in deploying LLMs within the healthcare domain, as discussed by Harrer et al. \cite{harrer2023attention}.

\noindent\textbf{Education.} The use of LLMs, e.g., ChatGPT, in education is associated with significant drawbacks, particularly in fostering inaccurate concept learning and an inappropriate approach to education. ChatGPT, being a language model trained on diverse Internet data, may unintentionally propagate misinformation or present concepts with a lack of precision and educational rigor, for instance, scientific misconduct \cite{mozes2023use}. The excessive dependence on LLMs by both educators and learners can have serious adverse effects. Students engaging with ChatGPT may encounter misleading content or promote misconceptions, potentially compromising the quality of their learning experience. The absence of real-time fact-checking and the model's susceptibility to biases and errors pose risks, potentially leading learners astray and impeding their overall educational progress. Consequently, caution is recommended when relying on ChatGPT as an educational tool without appropriate supervision and verification \cite{senechal2023balancing}, \cite{milano2023large}.

\noindent\textbf{Governance.} The potential misuse of LLMs in governance for spear phishing presents significant cybersecurity challenges. Using GPT-4 as an illustrative example, personal information of British members of parliament (MP) was extracted from Wikipedia, and GPT-3.5 was utilized to generate biographies, which were then incorporated into phishing emails sent to official email addresses \cite{hazell2023large}. This highlights the risks associated with misinformation, biased content, and the utilization of LLMs in AI-based cyberattacks within governance. The leakage of confidential information through such attacks can pose severe consequences for national security. The generation of misinformation and hate speech by LLMs further emphasizes the existing challenges, underscoring the imperative need for robust safeguards and countermeasures to address the risks related to the usage of these models in governance settings \cite{hazell2023large}.

\noindent\textbf{Science.} Hallucinations, biases, and paradigm shifts are pressing concerns of LLMs in the science domain. There is a risk of LLMs generating non-existent and false content. For instance, Meta developed an LLM named Galactica for reasoning scientific knowledge. That was reported to generate major flaws due to reproducing biases and presenting falsehoods \cite{metaG}. As a result, the model was shut down just after launching public access \cite{heaven2022meta}. 
Another concern lies in the involvement of LLMs in the scientific discovery process. It is challenging to interpret and understand LLMs due to their black-box nature, raising doubts about their reliability and trustworthiness in the science domain. For example, peer-review reports generated by LLMs may misinterpret research articles, which may impair the peer review quality~\cite{sci}.
Moreover, collaborating with LLMs won't be fundamentally the same as collaborating with other researchers or experts in a corresponding field \cite{binz2023should}. Clear principles of using these LLMs and/or other AI tools in scientific explorations should be established to ensure transparency, fairness, and trustworthiness \cite{binz2023should}.

\section{Limitations of Existing Works and Future Research Direction}
\begin{figure}
    \centering
    \includegraphics[width=\textwidth]{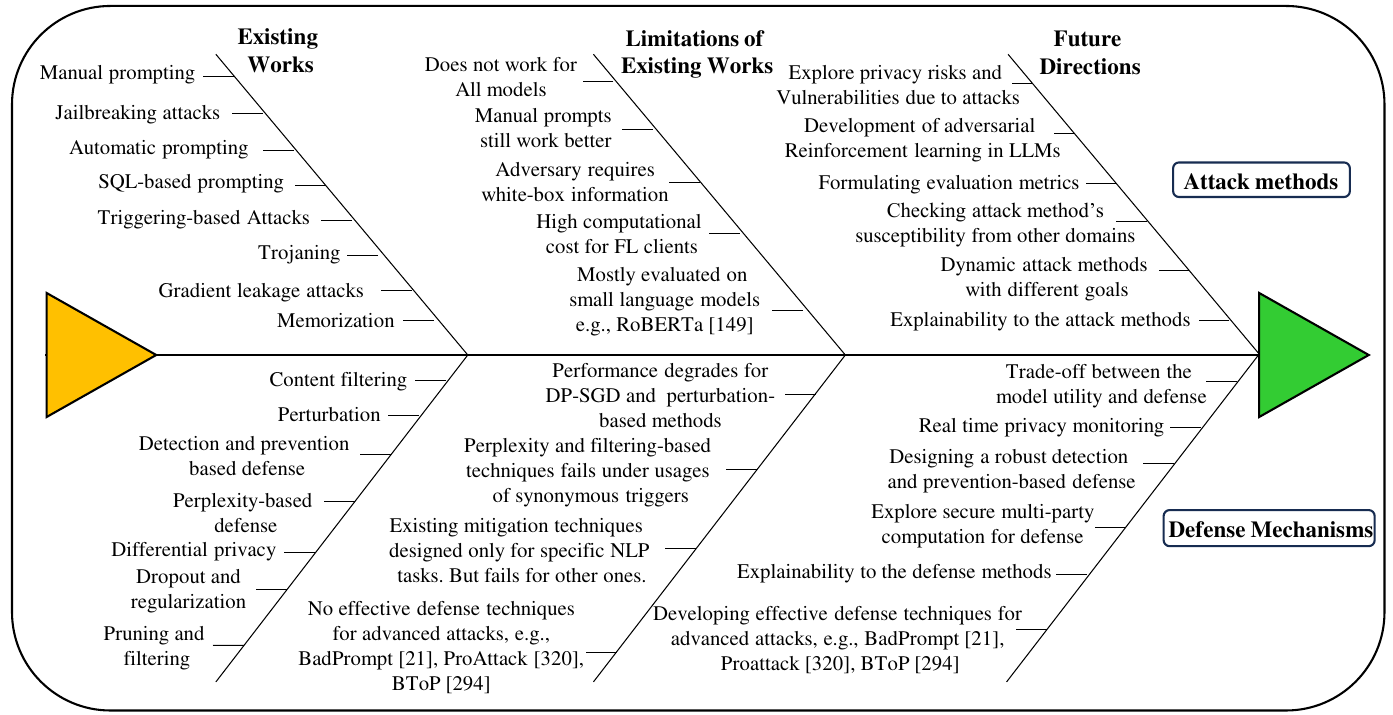}
    \caption{Overview of the advancements of the attack methods, defense mechanisms in LLMs, their limitations, and future research directions.}
    \label{fig:fishboneDiagram}
    \vspace{-2ex}
\end{figure}

Following a comprehensive examination of prevailing security and privacy attacks and defense mechanisms, this section delves into the prospects of advancing secure and privacy-preserving LLMs. In Figure~\ref{fig:fishboneDiagram}, we show an overview of the evolution of attack methods and defense mechanisms in LLMs, their limitations, and future research directions. We then 
discuss the limitations of current security and privacy attacks and defenses along various promising domains that require further research.

\noindent \textbf{Existing Attack Methods, their Limitations, and Future Research Direction: 
}
{\color{black}In LLMs, various categories of security and privacy attacks have emerged, posing significant risks to LLM systems. Among security attacks, prompt injection is a prominent technique in which attackers craft malicious prompts, either manually~\cite{liu2023prompt} or automatically~\cite{shin2020autoprompt}, to mislead the LLM into generating inappropriate or harmful outputs. These prompts are designed to bypass the model’s safety alignments~\cite{jiang2023prompt}, enabling the generation of content as per the attacker’s intent. Jailbreaking attacks involve creating malicious prompts in tricky way~\cite{spennemann2023exploring}, such as character role-play~\cite{gupta2023chatgpt} and attention shifting~\cite{liu2023jailbreaking}, so that the LLM generates inappropriate/harmful contents. Data poisoning attacks involve the insertion of malicious data samples during the model's training/fine-tuning phases~\cite{wan2023poisoning,kurita2020weight}, introducing biases or vulnerabilities that compromise the model's functionality. Similarly, backdoor attacks introduce hidden ``backdoors'' (via trojaning) during the training/fine-tuning phase~\cite{shi2023badgpt}, which are activated by the presence of backdoor triggering words in the prompts, making the LLM system vulnerable. LLM privacy attacks focus on extracting sensitive information such as private training/fine-tuning data, personal information, and even model components (e.g., gradients or model architecture). Techniques used in these attacks include gradient leakage attacks~\cite{deng2021tag}, MIA~\cite{shokri2017membership}, and PII leakage attacks~\cite{nakka2024pii}.}

Recent studies have shown that existing attack methods have some drawbacks. For instance, the DAN attack \cite{shen2023anything} builds on jailbreak prompts gathered over six months, extending from the inception of ChatGPT-related sources to May 2023. It is recognized that adversaries have the potential to persist in refining jailbreak prompts for specific objectives beyond the documented collection time-frame. 
This demonstrates
the dynamic nature of adversarial strategies, with the understanding that new, optimized prompts may emerge even after the initial data collection phase. 
Moreover, most of the existing studies on jailbreaking attacks are primarily focused on ChatGPT \cite{dengmasterkey, shen2023anything, li2023multi}. It remains unclear whether potential vulnerabilities exist in other LLMs, such as Vicuna \cite{chiang2023vicuna}, Bard \cite{manyika2023overview}, and Bing Chat \cite{bingchat}. For example, \textsc{MasterKey} \cite{dengmasterkey} may not achieve comparable performance on Vicuna as it does on ChatGPT.
For prompt-based attacks, manual prompts are often more effective than automated prompt-based attack methods. As automated attack methods are primarily designed for generalized tasks, they may not demonstrate same effectiveness for specific tasks as manually crafted prompts. The unlabeled and imbalanced real-world data further complicate the development of effective automated prompting-based attack methods
\cite{shin2020autoprompt}. For backdoor attacks, most of them are focused on classification or similar tasks \cite{cai2022badprompt}, e.g., sentiment analysis and opinion classification. More attention should be paid to the attacks that perform other NLP tasks, including question answering, text summarization, and language translation. 

The underlying philosophy behind privacy attacks lies in the correlation between the level of accessibility an adversary holds and its ability to extract sensitive information or exert control over target victim LLMs. More access leads to a broader potential for the adversary to recover sensitive data or influence the target LLM \cite{li2023privacy}. For instance, when the adversary has access only to the black-box model, the attacker might be able to leverage training data extraction attacks to recover a limited set of private data. However, if the adversary is granted white-box information such as gradients, the attacker can leverage this extra information to accurately recover more private training instances \cite{deng2021tag}. This expanded access can facilitate various privacy attacks, including attribute inference attacks, embedding inversion, and gradient leakage attacks. In gradient-based attacks, the adversary needs the white-box information of a model, which is sometimes impractical in real-world practice. Moreover, most of the existing privacy attacks are designed for vision models. A limited number of studies have reported gradient leakage attacks specifically on language models, e.g., the LAMP attack \cite{balunovic2022lamp}. In essence, the increased access empowers adversaries to perform more sophisticated and targeted privacy attacks against LLMs, potentially compromising sensitive information or gaining access to the internal model architecture. The early MIAs were based on the white-box access assumption \cite{shokri2017membership}, which is sometimes impractical in real-world deployment.
The evaluation dataset for some attacks, e.g., ProPILE was built solely from private information available in open-source datasets provided by major corporations, ensuring ethical data acquisition \cite{kim2024propile}. However, it is crucial to note that the heuristic data collection process might potentially lead to instances of bias, disassociation, or noise. This adds uncertainty and potential inaccuracies in the benchmark dataset, requiring attention when interpreting the results. 

However, most attack methods (e.g., BadGPT, BadPrompt, and Trojaning attacks) described in the existing studies are designed for only relatively small NLP models. Only a few are tested on LLMs (e.g., ProPILE, DAN, and JAILBREAKER). Also, the high cost of accessing commercialized LLMs, such as GPT-3.5 or upper versions, contributes to the lack of attack evaluations on LLMs. Besides, the in-depth vulnerability analysis in terms of privacy attacks and security issues is still lacking for LLMs. One of the reasons can be attributed to the limited number of performance evaluation metrics (e.g., perplexity) in the language domain to comprehensively evaluate attack and defense effectiveness. Also, in the FL environment, it requires very high computational power to train LLMs with such large datasets \cite{sutitanic}. It is an open research challenge to develop effective attack and robust defense methods along with the proper evaluation techniques for LLMs. The correctness of LLM-generated content has always been a major concern in this area of research. Due to the knowledge gaps, i.e., missing or outdated information might always be present in LLMs. The problem of hallucination in LLM has been investigated and evaluated
from the knowledge-gap perspective but is yet to be investigated from other perspectives such as safety, i.e., abstaining to generate harmful content \cite{SocraSynth}.

{\color{black} For LLMs, comprehensive exploration of their vulnerabilities under security and privacy attacks remains an essential area of study. Future research should examine the applicability of well-established attack methods from other domains, such as computer vision, to LLMs. In addition to existing methods, novel attack techniques should be developed to comprehensively inspect various vulnerabilities of LLMs, potentially targeting multiple objectives with a single attack approach. 
Moreover, special attention should be given to developing appropriate metrics for evaluating the impact of vulnerabilities related to security and privacy attacks in LLMs. Explainable AI (XAI) can play a vital role in this domain by increasing transparency and explainability within attack systems, allowing for a better understanding of LLM vulnerabilities. Developing XAI techniques to enhance the interpretability of LLM vulnerabilities is another essential research direction for advancing the security and privacy of LLMs.  }

\noindent \textbf{Existing Defense Mechanisms, Challenges, and Future Research Direction:} 
{\color{black}To mitigate vulnerabilities posed by security and privacy attacks, various defense techniques have been proposed in existing research. Instruction defense~\cite{Instruction_Defense}, paraphrasing~\cite{jain2023baseline}, and re-tokenization~\cite{jain2023baseline} methods are commonly employed to defend against prompt injection attacks. To defense jailbreaking attacks, SmoothLLM~\cite{robey2023smoothllm}, LLM Guard~\cite{LLM_guard}, and perplexity filtering~\cite{gonen2022demystifying} are widely used. Data curation techniques~\cite{yan2024llm}, including filtering poisoned content~\cite{wan2023poisoning} and detecting trigger-embedded inputs~\cite{chen2018detecting}, are widely used to mitigate data poisoning attacks and backdoor attacks. On the other hand, DP-SGD~\cite{abadi2016deep} is a widely adopted technique to defend against gradient leakage attacks, MIAs, and PII leakage attacks. In particular, noise perturbation~\cite{zhu2019deep} is a common strategy to defend against gradient leakage attacks. Various techniques are prevalent to mitigate MIA in LLMs, including model pruning~\cite{liu2018fine} and knowledge distillation~\cite{hinton2015distilling}. To prevent PII leakage attacks, training data curation~\cite{yan2024llm} and restricting attacker access to training data or model~\cite{inan2021training} serve as viable countermeasures.
}

For defense, studies reported that ChatGPT's safety protections are good enough to prevent single jailbreaking prompts however, it is still vulnerable to multi-step jailbreaking \cite{li2023multi}. Moreover, the new Bing AI chatbot \cite{bing} is more vulnerable to these direct prompts. System-mode self-reminder defense techniques are inspired by the human-like reasoning capabilities of LLMs \cite{xi2023defending}. The more discerning question regarding LLM reasoning processes with or without self-reminder remains unsolved. To acquire a comprehensive understanding of the reasoning processes of large neural networks, more in-depth investigation is highly required. Although the side effects of self-reminder have been explored on typical user queries across various NLP tasks, evaluating its effect on any type of user query poses a challenge, which makes it difficult to fully understand its impact on user experience \cite{wu2023defending}.

Considering the shortcomings mentioned, developing more flexible self-reminding systems and expert frameworks that improve safety, trustworthiness, and accountability in LLMs without compromising effectiveness can be a fundamental research challenge to protect LLMs from jailbreaking attacks. Furthermore, individuals with malicious intent are highly active in online forums, sharing and discussing new strategies. Frequently, they keep these exchanges private to evade detection. Consequently, it is essential to conduct further research and studies aims to identify and implement effective defense strategies to mitigate the risks posed by the latest jailbreaking attacks. 
Efficient strategies for defending against backdoor attacks in a black-box environment are still lacking \cite{kandpal2023backdoor}. Existing defense mechanisms  (\cite{sur2023tijo}, \cite{qi2020onion}, \cite{shao2021bddr}), for specific learning tasks in LMs are not evaluated for the other learning tasks like, text summarizing, and prompt-based learning. Moreover, it is found in the literature that prompt-based PLMs are highly susceptible to textual backdoor attacks \cite{xu2022exploring}, \cite{du2022ppt}. Addressing the challenge of textual backdoor attacks in prompt-based paradigms, particularly in the few-shot learning setting \cite{wang2020generalizing}, is another unresolved challenge. MDP (Masking-Differential Prompting) defense \cite{xi2023defending} faces challenges in various NLP tasks like paraphrasing and sentence similarity \cite{gao2020making}. Its performance under few-shot learning remains uncertain due to a lack of practical evaluation. 
While MDP has demonstrated strength against earlier backdoor attacks, it may not be effective against recently introduced attacks like BadPrompt \cite{cai2022badprompt} and BToP \cite{xu2022exploring}. Perplexity-based methods and filtering-based methods may not work well when attackers use synonymous trigger keys \cite{huang2023composite}. Furthermore, developing dynamic defense methods considering the above factors is a challenging future task. Currently, the predominant focus of investigation on backdoor attacks revolves around text classification in LLMs. However, a notable gap exists in the literature concerning investigations into backdoor attacks on various tasks for which LLMs find widespread application, e.g., text summarization and text generation \cite{yang2023comprehensive}. Understanding and addressing backdoor attacks in various tasks for which LLMs are employed is crucial for developing effective defense mechanisms and ensuring secure deployment of LLMs. While poisoning attacks on ML models have been investigated in the literature \cite{mooreFL2023}, there is not yet an effective solution for several attack methods, including ProAttack\cite{zhao2023prompt} and Badprompt\cite{cai2022badprompt}. Further research in diverse tasks and models can enhance the knowledge and understanding of the security impacts of LLMs, as well as facilitate the development of robust and trustworthy LLM systems. Defense techniques, such as dataset cleaning, and removing near duplicate poisoned samples and anomalies, sometimes slow down the model development process in order to defend against data poisoning attacks. Other defense methods, e.g., stopping training after certain epochs, achieve a moderate defense against poisoning attacks but degrade the model utility \cite{wan2023poisoning}. Gradient perturbation \cite{hu2023defenses} and DP-SGD-based methods \cite{abadi2016deep} are frequently used to defend against privacy attacks in LLMs. It can prevent the private training data from being leaked based on the parameter configurations at a small cost of model utility. Limiting the accessibility to the model and generating limited prediction results might be another option \cite{zanella2020analyzing}. Extensive research studies can obtain proper knowledge of to what extent algorithmic defenses such as differential privacy can prevent PII disclosure without compromising model utility. In the post-processing phase, for API-access models such as GPT-3, it is advisable to integrate a detection module that examines the output text to identify sensitive information. If sensitive content is detected, the system should either decline to provide an answer or apply masks to safeguard the sensitive information \cite{huang2022large}. Also, for image models, a recent study has demonstrated that adding a standard level of random noise into the gradient update might not always work well to prevent gradient leakage attacks on medical images \cite{das2023privacy}. Recently, an open language model (OLMo) has been introduced to provide open access to the data, code, and model \cite{soldaini2023dolma}. The main purpose of OLMo is to facilitate open research on language models. They performed PII filtering to remove it from the data, and they also provided a tool to remove PII data upon request. Though it followed existing practices to obscure the PII exposure and identify and remove toxic content, it did not explicitly discuss the impacts of various attacks outlined in this survey paper and how to mitigate those risks. Secure multi-party computation \cite{cramer2015secure} can be another way to defend against privacy attacks in LLMs, which can be explored in future research endeavors. Considering the above limitations of existing defense techniques in LLMs, developing a defense mechanism for these privacy attacks for LLMs would be an imperative task.

{\color{black}Future research initiatives for enhancing security and privacy in LLMs can be directed toward several key areas. An ideal defense method should be able to effectively achieve a balance between model utility and security \& privacy protection. The development of real-time privacy monitoring systems is essential to improve the resilience of privacy-preserving LLMs. This necessitates the exploration of robust detection techniques against various security and privacy attacks. Furthermore, a thorough evaluation of less-explored defense techniques, such as secure multi-party computation (SMPC), is necessary to assess their effectiveness against LLM vulnerabilities. Finally, leveraging the capabilities of XAI can improve transparency in LLM defense mechanisms.
}

\noindent\textbf{Role of explainable AI in Enhancing Security and Privacy of LLMs:} {\color{black}
LLM's ability to make decisions on various learning tasks is often criticized due to its black-box nature, e.g., non-interpretable weights. It makes it more challenging for the new practitioners and developers of this field, as it hinders their ability to clearly interpret and understand its application, particularly in critical cases. Explainable AI (XAI) can help to bridge this gap by developing methods to interpret and explain complex LLM systems, the decision-making process, and outputs \cite{cambria2024xai}. Conducting studies and research to build explainable methods is highly essential to make the usage of LLMs more reliable and trustworthy. XAI can provide transparent insights into the inference process and the dynamic weight assignment by the attention mechanism of the LLMs, which enhances interpretability by highlighting the most influential input features that contribute to its predictions. This explainability makes the model more trustworthy and ensures that its output is explainable. Additionally, it facilitates error analysis and bias detection. XAI has shown its potential in several real-world AI applications such as anomaly detection \cite{30dayscoding, ali2024next}, cyber-security \cite{saarela2024recent}, and enhancing data privacy \cite{ezzeddine2024privacy}. In ML, several popular methods exist such as \textbf{L}ocal \textbf{I}nterpretable \textbf{M}odel-agnostic \textbf{E}xplanations (LIME) \cite{ribeiro2016should}, and \textbf{SH}apley \textbf{A}dditive ex\textbf{Pl}anations (SHAP) \cite{lundberg2017unified}, which offer insights into model behavior without requiring access to internal model components. These methods analyze the association between inputs and outputs to identify features that most significantly contribute to model predictions. The explainability of code generation models such as CodeBERT \cite{feng2020codebert} and GraphCodeBERT \cite{guo2020graphcodebert} were proposed to understand code syntax and semantics \cite{yang2024robustness}. Also, the ad-hoc explanation further clarifies the model’s decision \cite{lipton2018mythos}. 

The role of XAI in the security and privacy aspect of LLM is crucial. Specifically in its development phase, having prior knowledge about the attackers' ability (e.g., black-box access to the model, injecting poisonous instances to the training/fine-tuning data) and the vulnerable components (training/fine-tuning data and the model itself) of LLM systems will assist in developing robust techniques against security and privacy attacks. However, XAI may expose critical aspects of the LLMs, including their architecture, components, and the dynamic weight allocation by attention mechanism that contribute to predictions. Such disclosures may increase the LLM's vulnerability to security and privacy attacks. For example, revealing the white-box nature of the model may facilitate the attacker's access to the model, increasing the risk of gradient leakage attacks or MIA \cite{ezzeddine2024privacy}. Recent studies have claimed that explainability in LLMs may raise further security concerns, especially with insidious backdoor attacks \cite{cheng2023backdoor}. Lin et al. proposed an XAI approach that can identify the triggers (backdoor attack) that mislead the model to contribute to error classification \cite{lin2021you}. Li et al. developed an XAI technique to identify and quantify the influence of raw data features on successful MIAs. This approach analyzes the data distribution to identify the influential neurons contributing to compromising private data and subsequently trains an MIA ensemble model using attack features derived from the selected neurons \cite{lin2021you}. XAI techniques in image models (e.g., Grad-CAM \cite{selvaraju2017grad}) may compromise model privacy under various attacks, such as gradient leakage attacks. In some cases, XAI enables more accurate reconstruction of private training data compared to models that rely solely on predictions \cite{zhao2021exploiting}. XAI-aware model extraction attack (XaMEA) was proposed to exploit spatial knowledge from decision explanations \cite{yan2023explanation}. 
It illustrated that the transparency provided by XAI may facilitate the attacker's access to the model, making it easier to exploit it. This increased explainability can make the model more vulnerable to model extraction attacks compared to prediction-only models. However, the vulnerabilities mentioned above are mostly explored for DNNs. In LLMs, the vulnerabilities in the XAI context have been explored to a limited extent. Moreover, there are several unique characteristics of LLMs that are different from DNNs, e.g., large-scale data and model parameters, task-agnostic, and semantic language understanding, which makes it more challenging to design XAI methods for better interpretations. The comprehensive interpretation of LLM vulnerabilities under XAI, including backdoor attacks, membership inference attacks (MIA), and model extraction attacks, has yet to be fully explored. Additionally, vulnerabilities related to prompt injection and jailbreaking attacks in the context of XAI still remain unexplored. Furthermore, open-sourced LLMs are publicly accessible and may provide greater explainability than closed-source models. However, this enhanced accessibility can make them more vulnerable to attacks when XAI techniques are employed. Thus, these LLMs may be more susceptible to attacks due to XAI. Further research is essential to fully understand the extent and impact of attackers' capabilities under XAI. Additionally, mitigation techniques should be developed in accordance with the identified risks and impacts.

 }

\section{Conclusion}
LLMs lend themselves as strong tools for comprehending complex linguistic patterns and generating logical and contextually coherent responses. However, such powerful models also entail potential privacy and security risks. In this survey, we first provided a detailed overview of LLMs' security and privacy challenges. We then discussed and analyzed the LLM vulnerabilities from both security and privacy aspects, existing mitigation and defense strategies against these security attacks and privacy attacks, as well as highlighted their strengths and limitations. In our investigation, we found that LLMs are highly vulnerable to the discussed attacks. According to our survey, there are a limited number of mitigation techniques to prevent those attacks against LLMs. The existing mitigation techniques that are applicable to relatively small LMs could potentially be used for LLMs. However, extensive research studies should be performed to evaluate and tailor the existing solutions to LLMs. Based on our analysis, we also outlined future research directions focusing on security and privacy aspects, pointed out key research gaps, and illustrated open research problems. The overarching goal is to enhance the reliability and utility of LLMs through comprehensive exploration and resolution of these vulnerabilities and offer pathways for future research toward secure and privacy-preserving LLM systems.

\noindent \textbf{ACKNOWLEDGEMENTS:} This work is partially supported by the U.S. Department of Homeland Security Grant Award Number 17STCIN00001-05-00. Further, M. Hadi Amini's work is partly supported by the U.S. Department of Homeland Security under Grant Award Number 23STSLA00016-01-00. 
The views and conclusions contained in this document are those of the authors and should not be interpreted as necessarily representing the official policies, either expressed or implied, of the U.S. Department of Homeland Security.

\bibliographystyle{ACM-Reference-Format}

\bibliography{sample-base.bib}

\appendix

\end{document}